%% file: main.tex
\pdfoutput=1

\documentclass[11pt]{article}

\usepackage[]{EMNLP2023}

\usepackage{times}
\usepackage{latexsym}

\usepackage[T1]{fontenc}

\usepackage[utf8]{inputenc}

\usepackage{microtype}

\usepackage{inconsolata}
\usepackage{graphicx}
\usepackage{array}
\usepackage{multicol,multirow}
\usepackage{makecell}
\usepackage{booktabs}
\usepackage[T1]{fontenc}
\usepackage{nicefrac}
\usepackage{pifont}
\usepackage{amsmath}
\usepackage{float}
\usepackage{enumitem}

\title{Bora: Biomedical Generalist Video Generation Model}

\author{
Weixiang Sun$^{1*}$, Xiaocao You$^{2*}$, Ruizhe Zheng$^{3*}$, Zhengqing Yuan$^{4}$, \\
\textbf{Xiang Li$^5$, Lifang He$^6$, Quanzheng Li$^5$, Lichao Sun$^6$}\\ 
    $^1$Northeastern University, China, $^2$Shanghai University of Finance and Economics\\ 
      $^3$Fudan University, $^4$University of Notre Dame\\
      $^5$Massachusetts General Hospital and Harvard Medical School, $^6$Lehigh University
      \\
  }

\begin{document}
\maketitle
\begin{abstract}
Generative models hold promise for revolutionizing medical education, robot-assisted surgery, and data augmentation for medical AI development. Diffusion models can now generate realistic images from text prompts, while recent advancements have demonstrated their ability to create diverse, high-quality videos. However, these models often struggle with generating accurate representations of medical procedures and detailed anatomical structures. This paper introduces Bora, the first spatio-temporal diffusion probabilistic model designed for text-guided biomedical video generation. Bora leverages Transformer architecture and is pre-trained on general-purpose video generation tasks. It is fine-tuned through model alignment and instruction tuning using a newly established medical video corpus, which includes paired text-video data from various biomedical fields. To the best of our knowledge, this is the first attempt to establish such a comprehensive annotated biomedical video dataset. Bora is capable of generating high-quality video data across four distinct biomedical domains, adhering to medical expert standards and demonstrating consistency and diversity. This generalist video generative model holds significant potential for enhancing medical consultation and decision-making, particularly in resource-limited settings. Additionally, Bora could pave the way for immersive medical training and procedure planning. Extensive experiments on distinct medical modalities such as endoscopy, ultrasound, MRI, and cell tracking validate the effectiveness of our model in understanding biomedical instructions and its superior performance across subjects compared to state-of-the-art generation models. Our model and codes are available at \url{https://weixiang-sun.github.io/Bora/}

\end{abstract}

\begin{figure}[t]
    \centering
    \includegraphics[width=\linewidth]{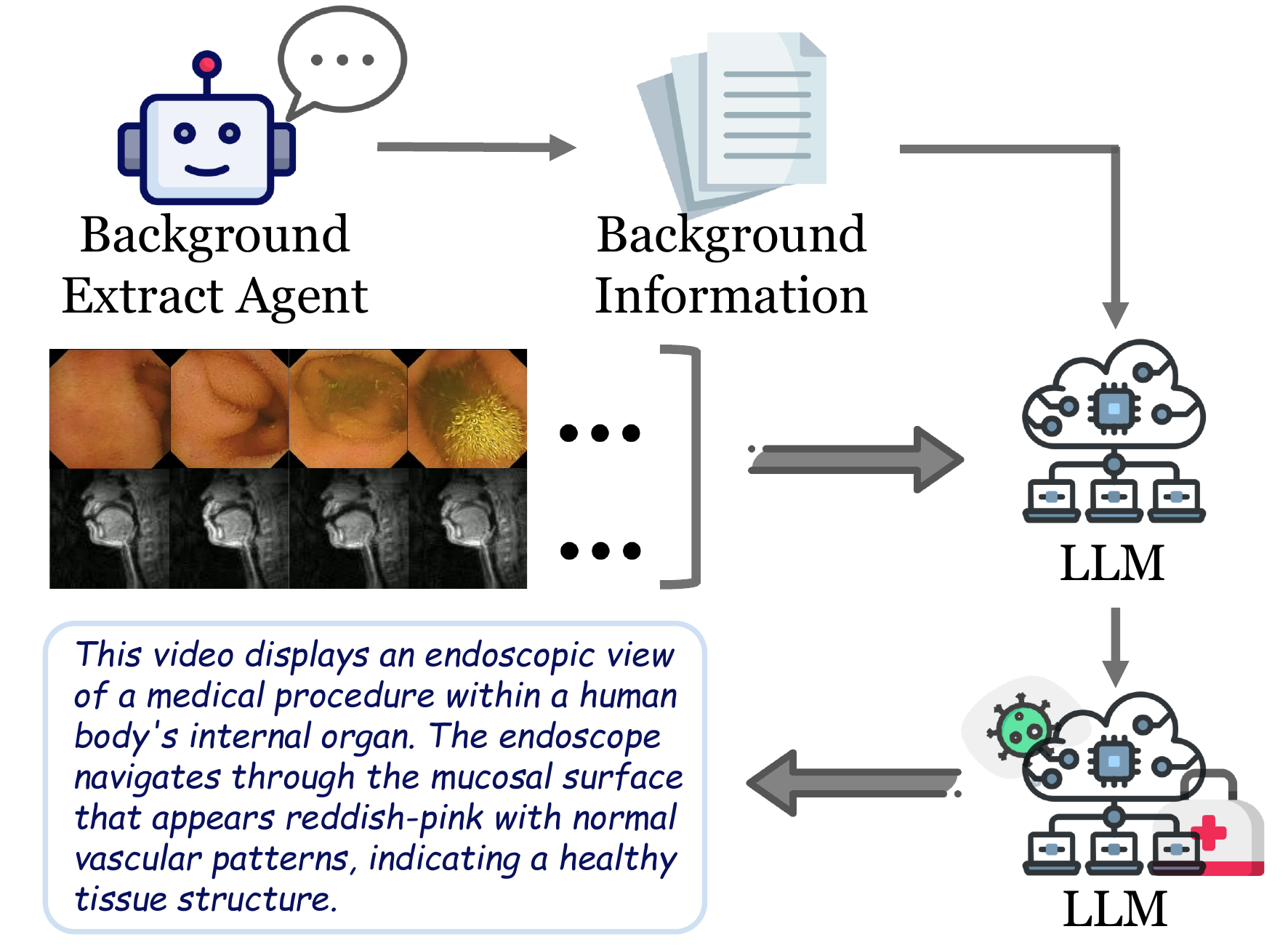}
    \caption{The overall process for generating captions. First, the agent extracts background information from the corresponding dataset, which is then injected into the LLM. Then, combined with the frame sequences, it generates high-quality captions.}
    \label{fig: wordcloud}
\end{figure}

\section{Introduction}

\begin{figure*}[!t]
    \centering
    \includegraphics[width = \linewidth]{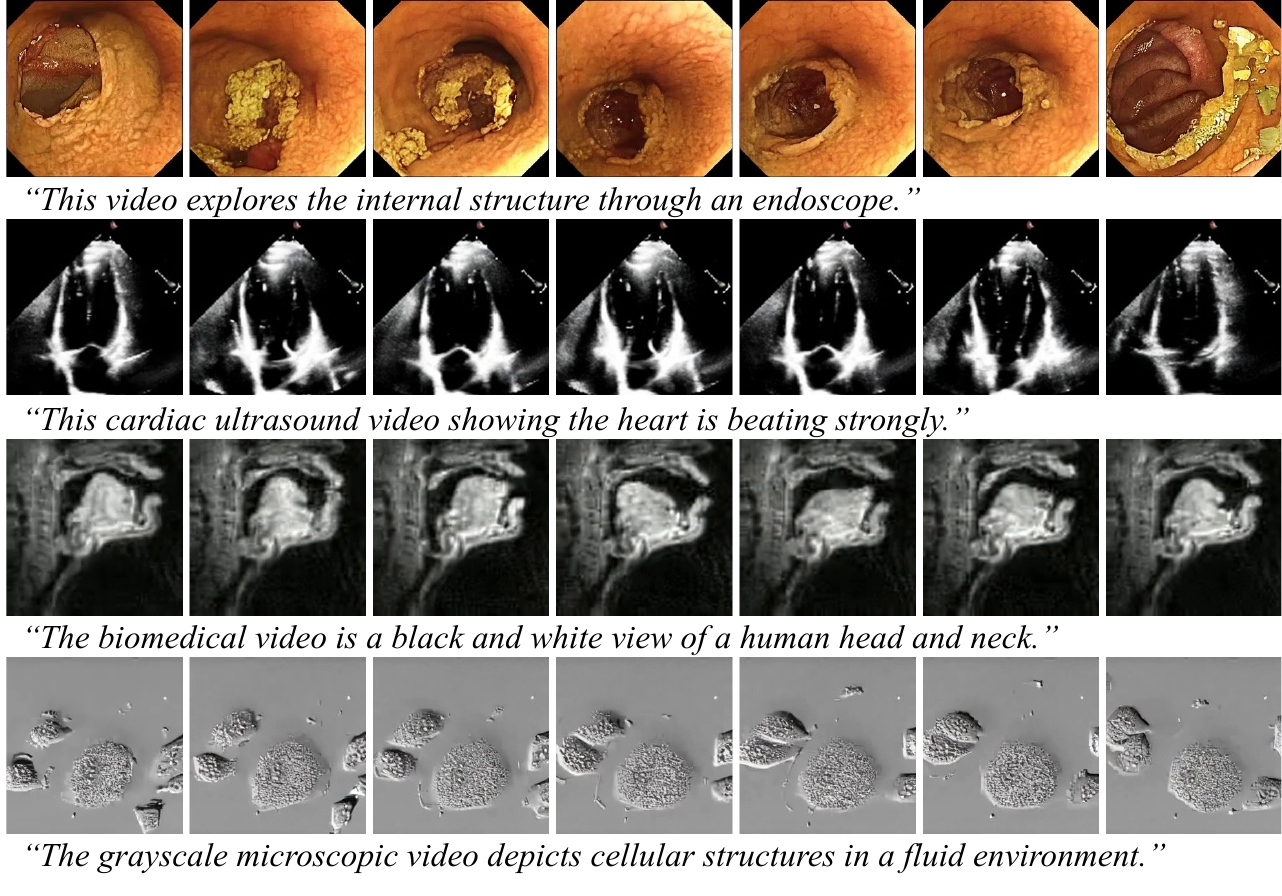}
    \caption{Some simple video examples produced by \textbf{Bora} and their corresponding text prompts showcase four biological modalities: endoscopy, ultrasound, real-time MRI, and cellular visualization.}
    \label{fig: sample}
\end{figure*}

Generative AI technologies have stepped into a new era, fundamentally altering how industries operate and deeply influencing everyday life~\cite{GPT4report}. Text-to-image (T2I) diffusion models are now capable of generating realistic images that adhere to complex text prompts~\cite{saharia2022photorealistic, rombach2022high}. Recent video generation models like Pika~\cite{pika}, SVD~\cite{SVD} and Gen-2~\cite{Gen-2} have also demonstrated their ability to create diverse and high-quality videos, primarily in general contexts. \textit{Sora}~\cite{liu2024sora}, introduced by OpenAI in February 2024 and known for its advanced capabilities in generating detailed videos from textual descriptions. It stands out by its capacity to generate high-quality videos lasting up to one minute, while faithfully following the textual instructions provided by the users~\cite{openai2024sora}. Despite their revolutionary capabilities, these models often struggle with generating accurate representations of medical procedures, detailed anatomical structures, and other specific clinical information.

So far, a diversity of models have been proposed for text-guided visual contents. Among them, diffusion models stand out as the most powerful deep generative models for various tasks in content creation, including image-to-image, text-to-image and text-to-video generation~\cite{croitoru2023diffusion,wang2023lavie,mogavi2024sora,ma2024latte}. Video generation aims to produce realistic videos that exhibit a high-quality visual appearance and consistent motion simultaneously. Diffusion models demonstrate strong performance in synthesizing temporally coherent video frames with flexible conditioning and controls, stronger diversity and more significant details. Finetuning of diffusion models is more cost-effective due to its generalizability and adaptability to user requirements. In particular, recent advance in Transformer-based large-scale video diffusion models has enabled long video generation in adherence to specific human instructions. \textit{Sora}\cite{liu2024sora} demonstrates a remarkable ability to accurately interpret and execute complex human instructions. 

However, there have been few attempts to explore biomedical video generation, which demands that the model comprehends complex medical instructions and intricate real-world dynamics. In addition, current approaches for video generation are not capable of generating accurate and realistic anatomical structures. In this work, we investigate the potential of diffusion models to generate biomedical video content with high controllability and quality. We begin by establishing a medical video corpus, which includes paired text-video data from various biomedical fields, encompassing both non-task-oriented and task-oriented content. Due to concerns regarding intellectual property and privacy, the dataset list is not exhaustive. Nevertheless, our text-video corpus is designed to be representative of diverse applications, ranging from macroscopic to microscopic scales. For video clips lacking consistent descriptions, we leverage LLM to generate captions, thereby enhancing the usability of their content. Then, we design Bora, the first spatio-temporal diffusion probabilistic model for text-guided generalist biomedical video generation. Bora is based on Transformer architecture and has been pre-trained on general-purpose video generation tasks. As shown in Figure 1, We fine-tune the model through alignment and instruction tuning on the constructed corpus. We assess whether Bora-generated videos appear plausible with respect to medical expert standards and evaluate their consistency and diversity. 

For generative video modeling, it is well-established that pre-training on a large and diverse dataset followed by fine-tuning on a smaller, higher-quality dataset is beneficial for final performance. Therefore, Bora is initialized from a pre-trained weights on large scale data. The model is fine-tuned through alignment and instruction tuning on the well-curated biomedical video corpus. Through extensive text-to-video generation experiments, we demonstrate that Bora is capable of generating high-quality video data with assistance from LLM-based captions across four distinct biomedical domains. More importantly, we assess whether Bora-generated videos appear plausible with respect to medical expert standards and evaluate their consistency and diversity. The results show that Bora achieves significantly better understanding of domain-specific instructions than general-purpose state-of-the-art video diffusion models. It also show promising subject and motion consistency across various modalities such as endoscope, RT-MRI and ultrasound imagings.

The debut of Bora, a generalist biomedical video generative model, underscores its vast potential in enhancing medical consultation, diagnosis, and operations for clinical practitioners, thereby improving patient experience and welfare. Bora can significantly impact medicine by providing patients with visual guides on procedures and treatments and offering doctors real-time assistance. In medical education, Bora could offer resources for students. Additionally, Bora could accelerate the integration and development of AR/VR technologies for immersive medical training and procedure planning. We summarize the contributions as follows:

\begin{itemize}[nolistsep, leftmargin=*, topsep=0pt]
    \item We propose Bora, a generalist biomedical video generation model. Extensive experiments highlight Bora's superior performance over other models in terms of video quality and consistency and its capability in following expert instructions.
    \item Given the limited availability of high-quality data, we construct the first comprehensive biomedical video-text corpus by extracting detailed descriptions and background knowledge from open-source video data using LLM. This is expected to provide valuable resources for future research.
    \item We validate Bora's capability in generating videos across various biomedical modalities, including endoscopy, ultrasound, real-time MRI, and cellular motility. Bora's proficiency in producing diverse realistic medical videos opens new avenues for medical AI.
\end{itemize}

\section{Related Work}
\noindent\textbf{Text-to-Image Diffusion Model} So far, most of the SOTA approaches for text-to-image generation are based on diffusion models\cite{achiam2023gpt,saharia2022photorealistic}. Diffusion models constitute a class of generative models that utilize diffusion stochastic process to modeling data generation. It can be conditioned by the class-induced and non-class-induced information\cite{ho2022classifier}, while the latter has become the predominant approach due to its flexibility. Of these, DALL-E2\cite{ramesh2022hierarchical} and Imagen \cite{saharia2022photorealistic} achieve photorealistic text-to-image generation using cascaded diffusion models, whereas Stable Diffusion \cite{rombach2022high} performs generation in a low-dimensional latent space.

\noindent\textbf{Text-to-Video Diffusion Model} Recent years have witnessed significant discussion on video generative models. Text-to-video extends text-to-image generation to generation of coherent high-fidelity videos given text conditions. At the initial phase, Variational Autoencoders (VAEs) and Generative Adversarial Networks (GANs)~\cite{li2017video, tulyakov2017mocogan, Chu_2020, yaohui2020imaginator} are used to model spatio-temporal dynamics of video data, which is vulnerable to mode collapse. Diffusion models, by contrast, can generate dynamic and accurate video content with improved stability~\cite{ho2022imagen, singer2022makeavideo}. Specifically, spatial and temporal modules are leveraged to generate time-consistent contents. For example, MagicVideo~\cite{zhou2023magicvideo} introduces latent diffusion to the text-to-video generation framework, enhancing model's capabilities to capture video content dynamics. \cite{ge2023preserve} proposes a video noise prior to boost performance. Several solutions to the high computational cost of training video diffusion models have been proposed, such as downsampling the videos spatio-temporally~\cite{bar2024lumiere} or fine-tuning only the temporal modules~\cite{blattmann2023stable, guo2023animatediff} by reusing pre-trained weights.

\input{weixiang/tab/data}

\textbf{Diffusion Models in Biomedicine}
In recent years, the application of generative models has expanded significantly, evolving from conventional domains to specialized industries, including the field of biomedicine. So far, several works have applied diffusion models in synthesizing medical visual contents of a variety of modalities for data augmentation and privacy protection in development of AI models for medical image analysis. \cite{xu2023medsyn} uses text-conditioned synthesized low-resolution images as a foundation for 3D CT images.\cite{dorjsembe2022three} proposes the first diffusion-based 3D medical image generation model that achieves promising results in high-resolution MRI image synthesis. However, biomedical video generation is yet to be explored. \cite{endora} proposes \textit{Endora}, a preliminary attempt to develop a video diffusion model specifically for endoscope data. To the best of our knowledge, there has been no medicine-specific generative model for producing high-quality and accurate videos.

\input{weixiang/dataset}
\input{weixiang/bora}
\input{weixiang/exp}
\section{Conclusion}
In this work, we propose Bora, a model designed for generating high-quality biomedical videos. By combining detailed descriptions and extensive background knowledge from video data, we created the first high-quality biomedical text-video pair datasets, highlighting the importance of open-source data in medical AI. Bora sets a new standard with state-of-the-art accuracy and authenticity, surpassing other video generation models in understanding real-world scenarios. Its flexibility in video synthesis makes Bora valuable for various medical applications. We believe that our work will significantly advance subsequent developments in biomedical generation as well as in industries such as AR, VR, and even education.

\input{weixiang/limitation}

\bibliography{main}
\bibliographystyle{acl_natbib}

\clearpage

\appendix

\section{More Dataset Construction Details}
\label{apx: data}
Our preprocessing approach for video data primarily consists of two steps: resolution normalization and sampling. For resolution, we standardize all video data to a resolution of 256x256 pixels. The majority of the videos have an aspect ratio of 1:1 or close to it. For videos with unusual aspect ratios, given the characteristics of medical videos where crucial information is typically centered, we first compress or expand the shortest dimension to 256 pixels and then apply center cropping to achieve the desired resolution. Regarding sampling, to train models that can generate videos with pronounced movements, we set a threshold \( K \) to increase the frame interval during sampling. This method not only ensures a dynamic quality in the training data but also augments the dataset volume.

\begin{figure}[!t]
    \centering
    \includegraphics[width=\linewidth]{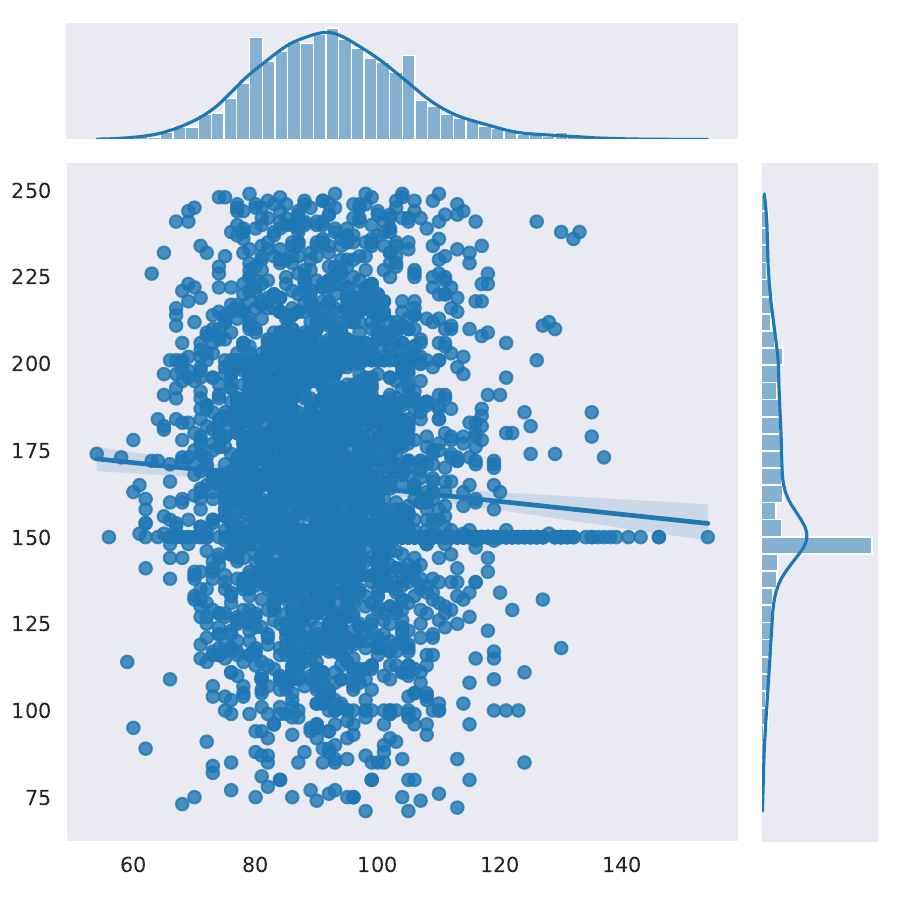}
    \caption{The distribution of video length (on the y-axis) and caption length (on the x-axis) in our text-video pair dataset, along with its fitted curve.}
    \label{fig: dis}
\end{figure}

In addition, we conducted simple statistics in conjunction with the captions we generated. The results depicted in Figure~\ref{fig: dis} illustrate the relationship between the number of video frames and the length of the captions. The video duration is largely concentrated around 150 frames, which is due to our model's objective to generate five-second videos. However, there is a distribution across other durations as well, allowing the videos to learn more temporal information for more accurate output. Regarding caption length, it generally presents an average of around 95 characters, with almost no captions shorter than 60 characters. This also indirectly reflects the consistency of the captions generated by our system.

\section{More Implementation Details}
\label{apx: impl}
\subsection{Training Details}
Our training process was conducted on four A100-80G GPUs. We accelerated the process by setting the data format to bf16, incorporating gradient checkpointing, and utilizing ZeRO-2 optimization. Specifically, we set the batch size to 4, the learning rate to \(1 \times 10^{-5}\), and the gradient clipping threshold to 1.0. Besides these, we will specifically focus on the selection of pre-trained models within our architecture.

\noindent\textbf{Text Encoder Selection. }
To obtain representations in the embedding space of text, one can either utilize a text-image encoder or a standard language model. Language models are trained solely on textual corpora, which are substantially larger than paired image-text datasets, thereby exposing them to a rich and diverse distribution of text. These models are typically much larger than the text encoders used in current image-text models. In this paper, we opt for the T5~\cite{raffel2020exploring} model from the language model category. T5 retains most of the original Transformer architecture, featuring sequence-independent self-attention that uses dot products instead of recursion to explore relationships between each word and all other words in a sequence. Positional encodings are added to the word embeddings prior to dot products; unlike the original Transformer, which uses sinusoidal positional encodings, T5 employs relative positional embeddings. In T5, positional encoding relies on the extension of self-attention to compare pairwise relationships, with shared positional encodings that are reevaluated across all layers of the model. As noted in Imagen, T5 demonstrates significant advantages in alignment and fidelity over image-text models such as CLIP. Therefore, we have reason to believe that this large-scale model, even without training in medical terminology, is sufficient as an encoder for encoding medical prompts.

\noindent\textbf{Video Encoder Selection. }
Sora employs a spatiotemporal VAE to reduce the temporal dimension. However, there are no high-quality spatiotemporal VAEs available in open source. Additionally, Open-Sora has indicated that the 2x4x4 VAE of VideoGPT is of low quality. Consequently, we have opted to use a 2D VAE from Stability-AI.

\onecolumn

\section{More Bora Samples and Compare with other models}
\subsection{Endoscopy}

\begin{figure*}[!h]
    \centering
    \includegraphics[width=\linewidth]{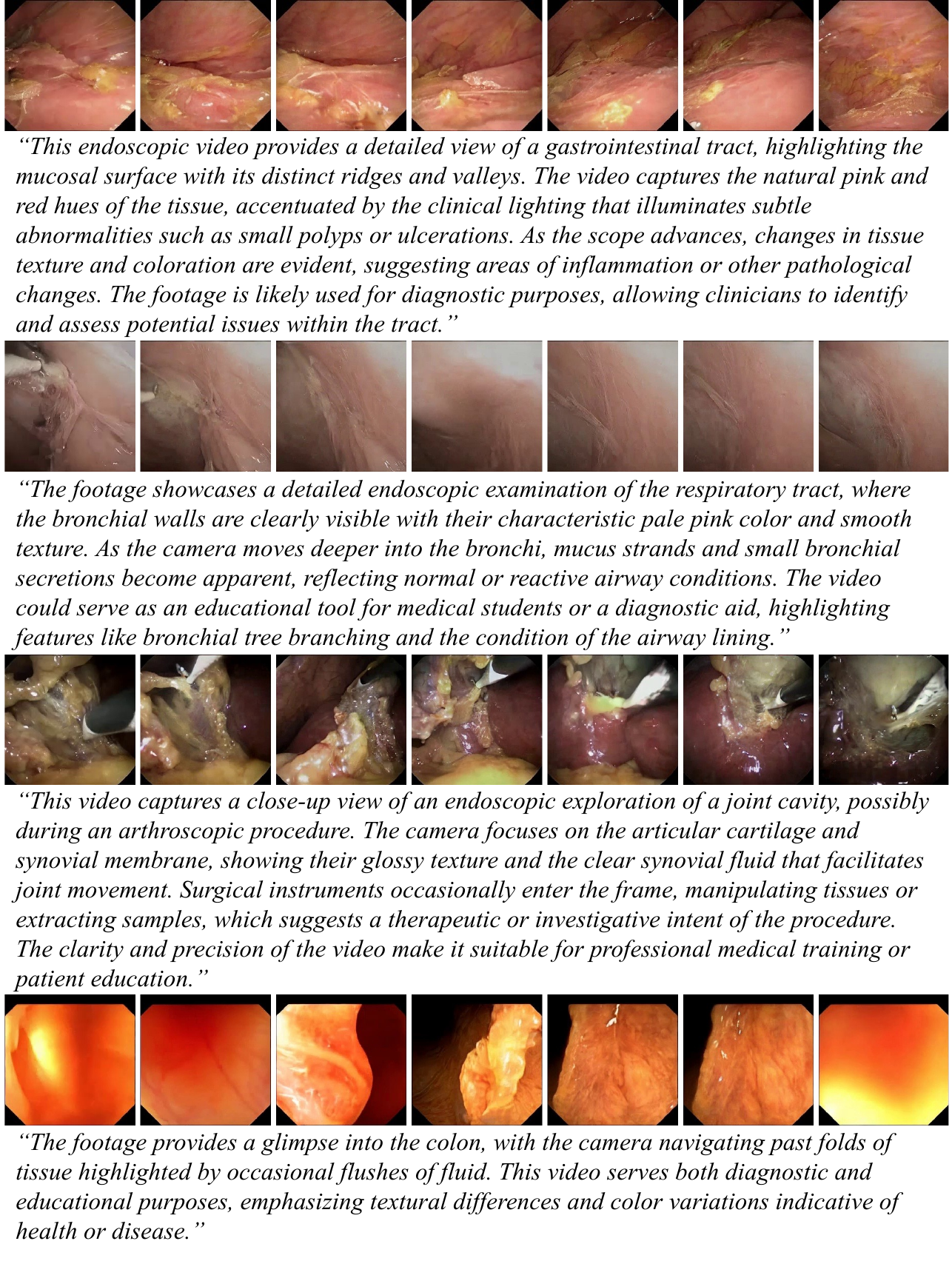}
    \caption{More sample generated by Bora in endoscopy modality.}
    \label{fig: endo show}
\end{figure*}

\clearpage

\subsection{Ultrasound}

\begin{figure*}[!h]
    \centering
    \includegraphics[width=\linewidth]{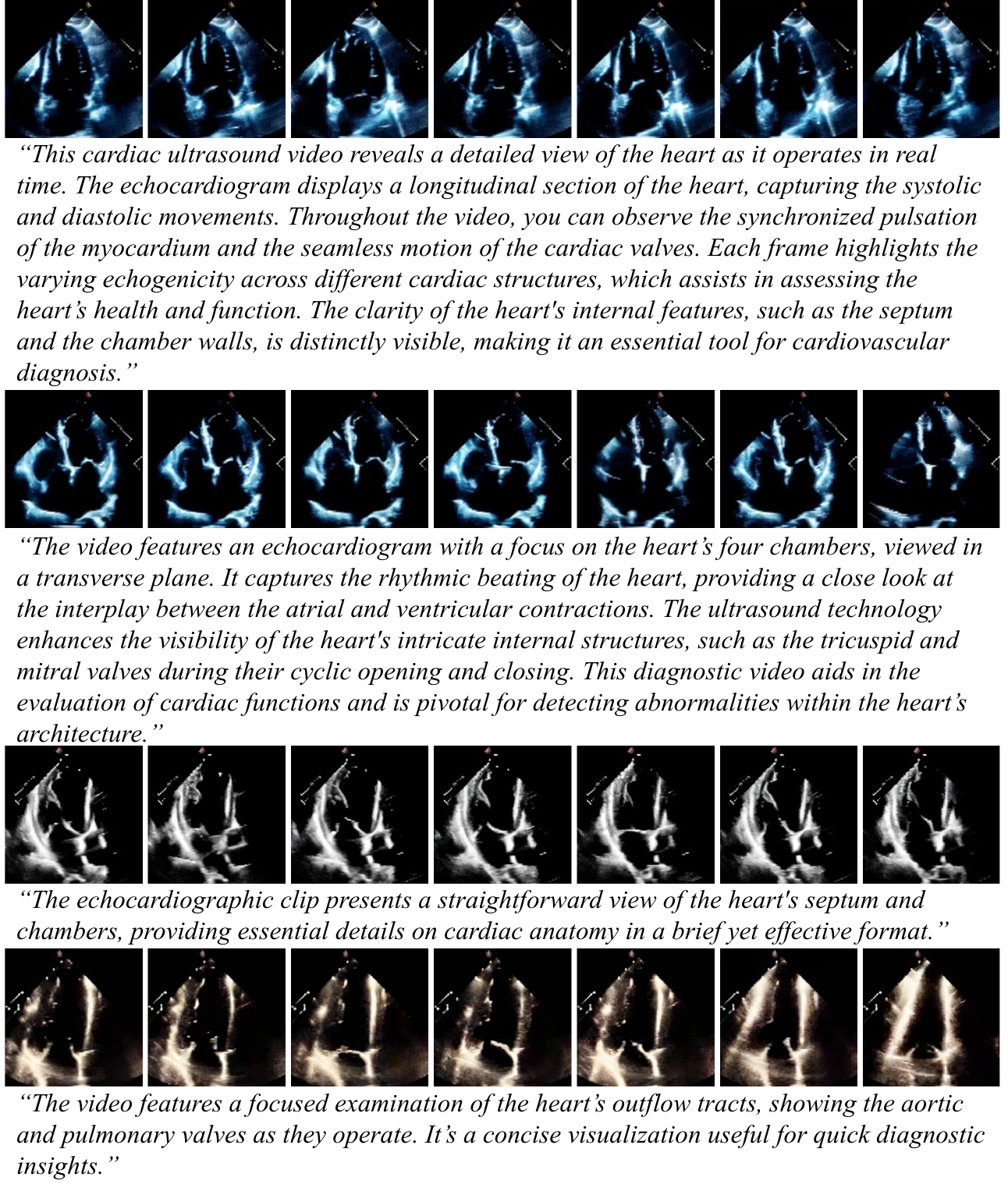}
    \caption{More sample generated by Bora in ultrasound modality.}
    \label{fig: uls show}
\end{figure*}
\clearpage

\subsection{RT-MRI}
\begin{figure*}[!h]
    \centering
    \includegraphics[width=\linewidth]{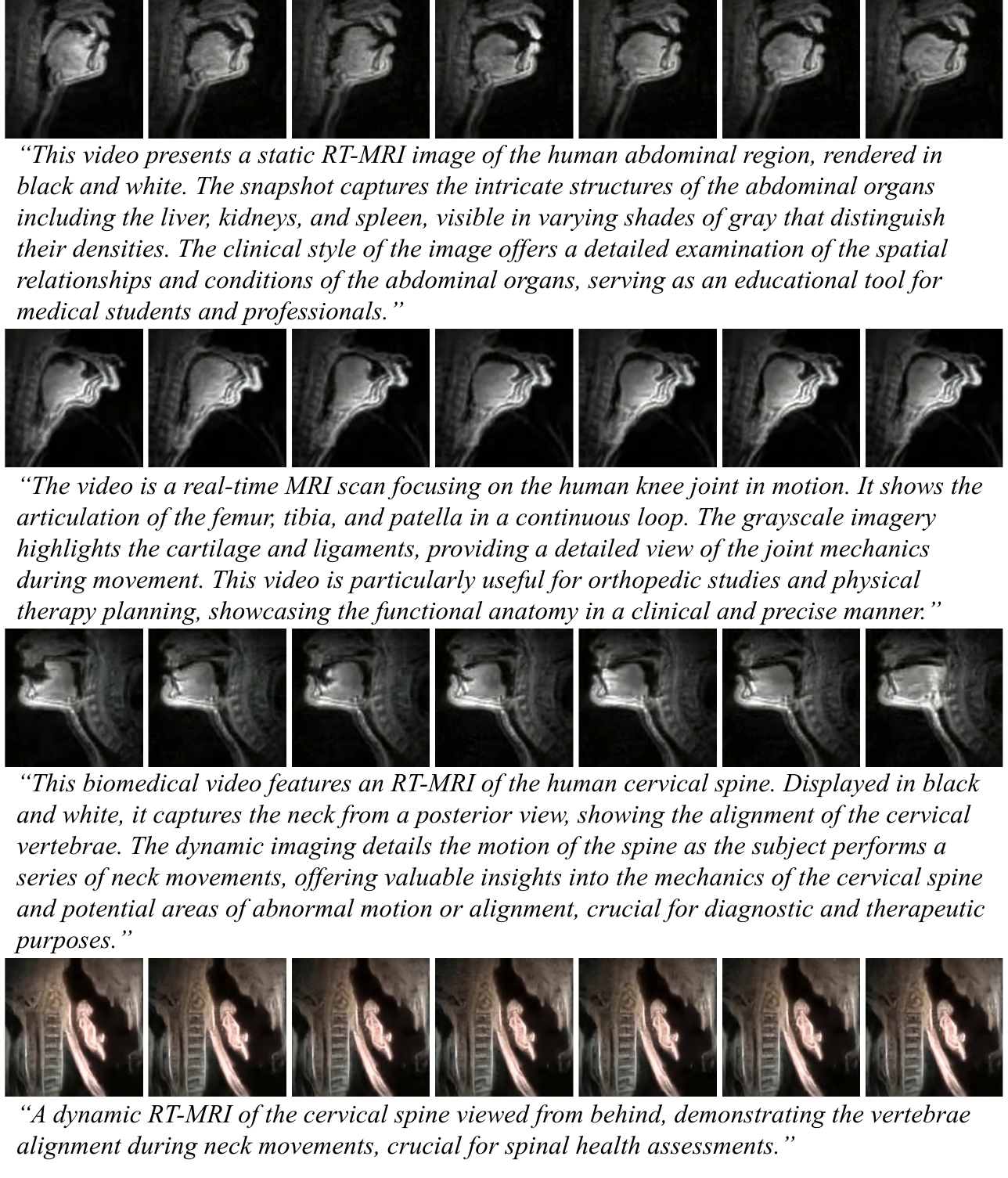}
    \caption{More sample generated by Bora in RT-MRI modality.}
    \label{fig: mri show}
\end{figure*}
\clearpage

\subsection{Cell}
\begin{figure*}[!h]
    \centering
    \includegraphics[width=\linewidth]{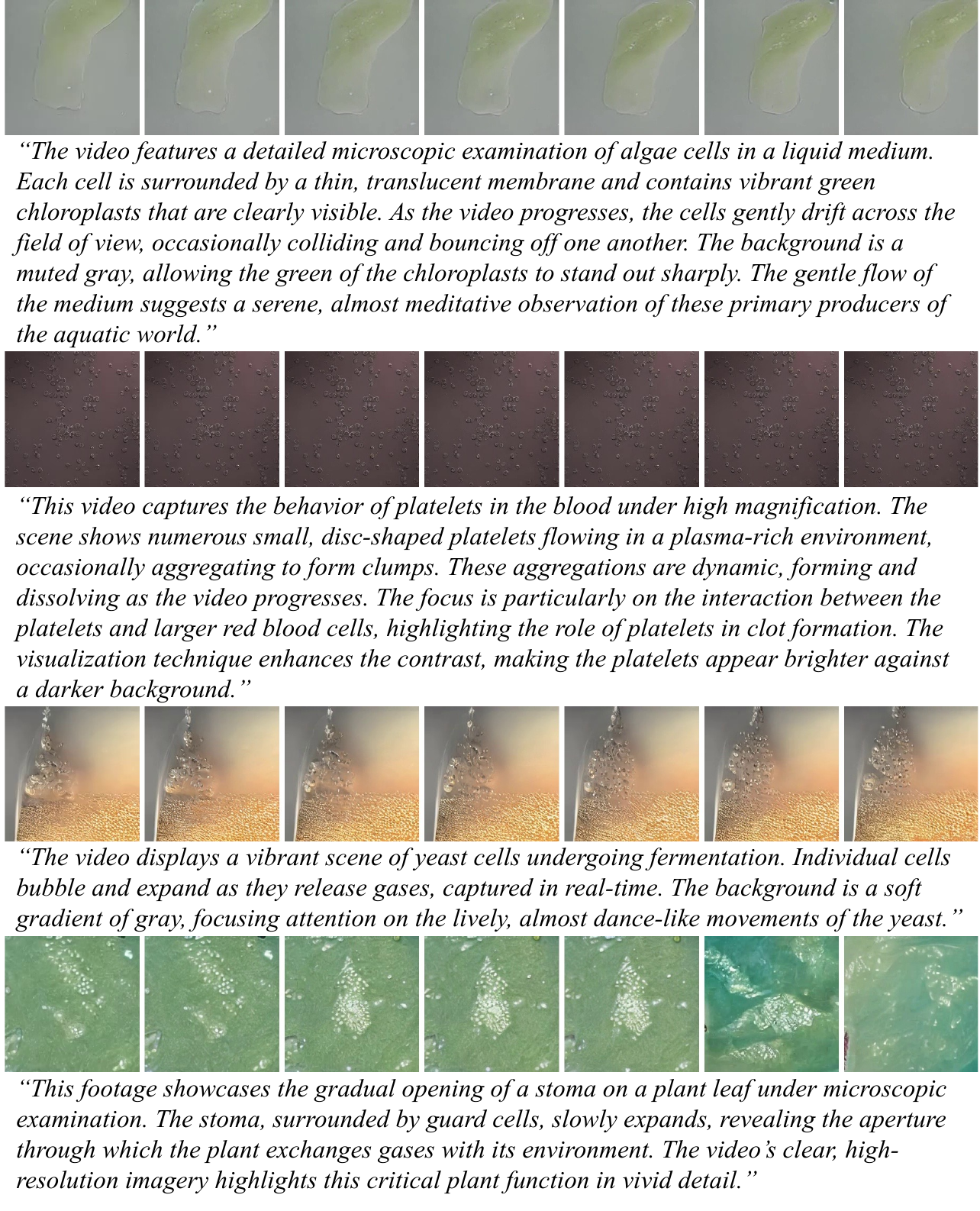}
    \caption{More sample generated by Bora in cell modality.}
    \label{fig: cell show}
\end{figure*}
\clearpage

\section{Bora \textit{vs} Others}
\subsection{Bora \textit{vs} Pika}

\begin{figure*}[!h]
    \centering
    \includegraphics[width=\linewidth]{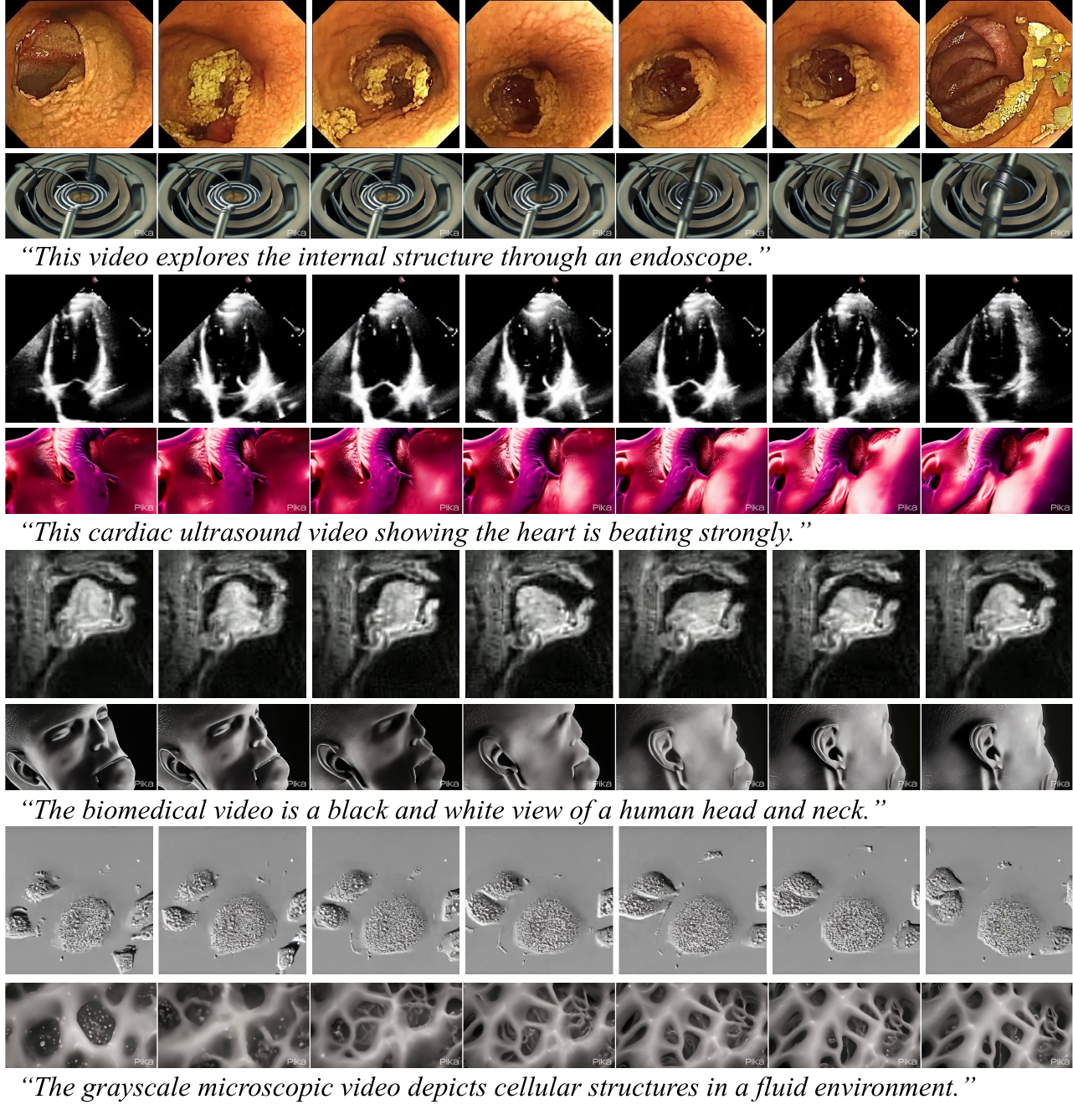}
    \caption{The comparison of simple prompts in four modalities between Bora and Pika}
    \label{fig: pika}
\end{figure*}

\clearpage
\subsection{Bora \textit{vs} PixVerse}
\begin{figure*}[!h]
    \centering
    \includegraphics[width=\linewidth]{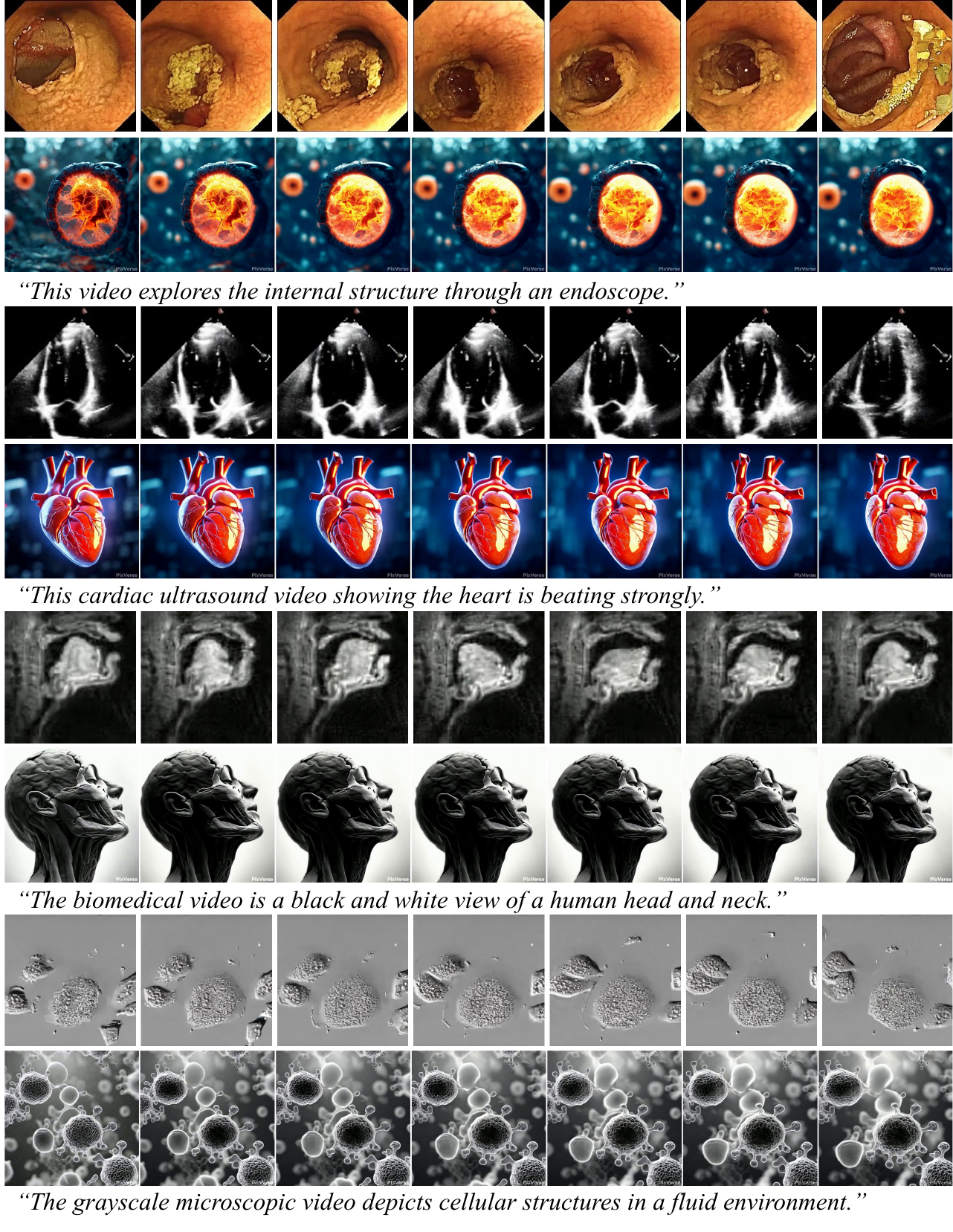}
    \caption{The comparison of simple prompts in four modalities between Bora and PixVerse}
    \label{fig: pix}
\end{figure*}
\clearpage

\subsection{Bora \textit{vs} Gen-2}
\begin{figure*}[!h]
    \centering
    \includegraphics[width=\linewidth]{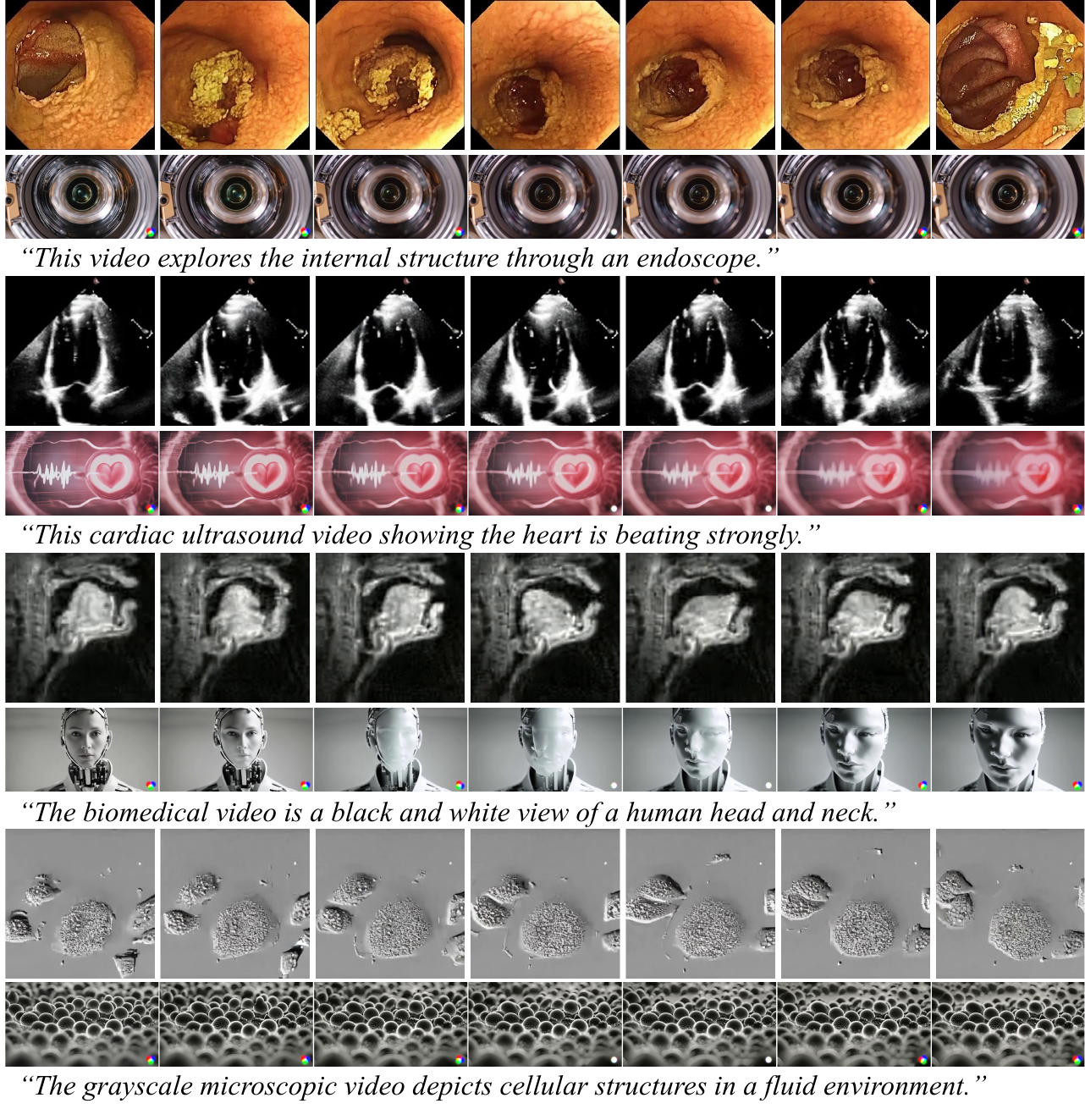}
    \caption{The comparison of simple prompts in four modalities between Bora and Gen-2}
    \label{fig: gen2}
\end{figure*}
\clearpage

\subsection{Bora \textit{vs} Model Scope}

\begin{figure*}[!h]
    \centering
    \includegraphics[width=\linewidth]{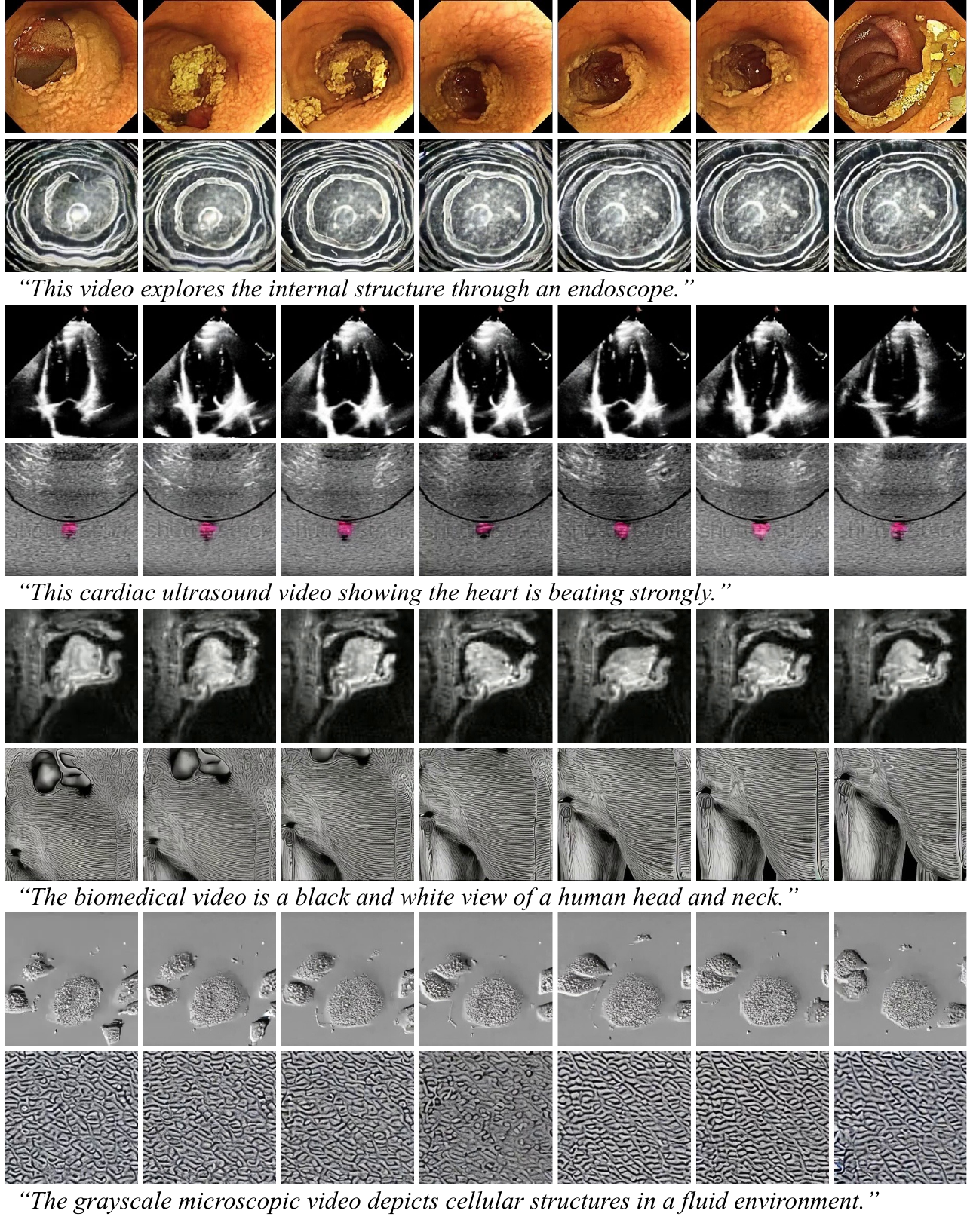}
    \caption{The comparison of simple prompts in four modalities between Bora and Model Scope}
    \label{fig: ms}
\end{figure*}
\clearpage

\subsection{Bora \textit{vs} Lavie}

\begin{figure*}[!h]
    \centering
    \includegraphics[width=\linewidth]{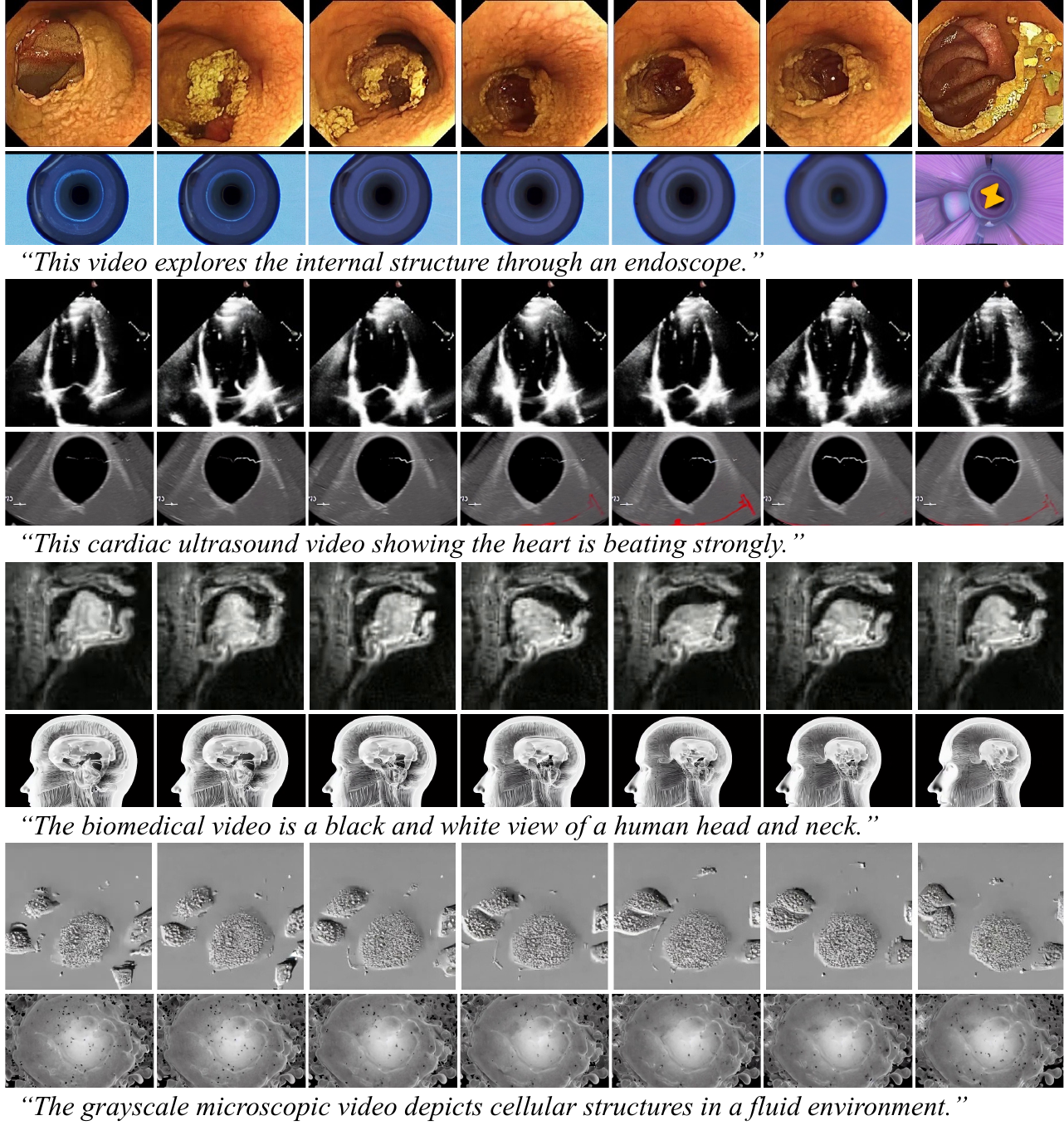}
    \caption{The comparison of simple prompts in four modalities between Bora and Lavie}
    \label{fig: lavie}
\end{figure*}

\section{Faliure Examples}
\begin{figure*}[t]
    \centering
    \includegraphics[width=\linewidth]{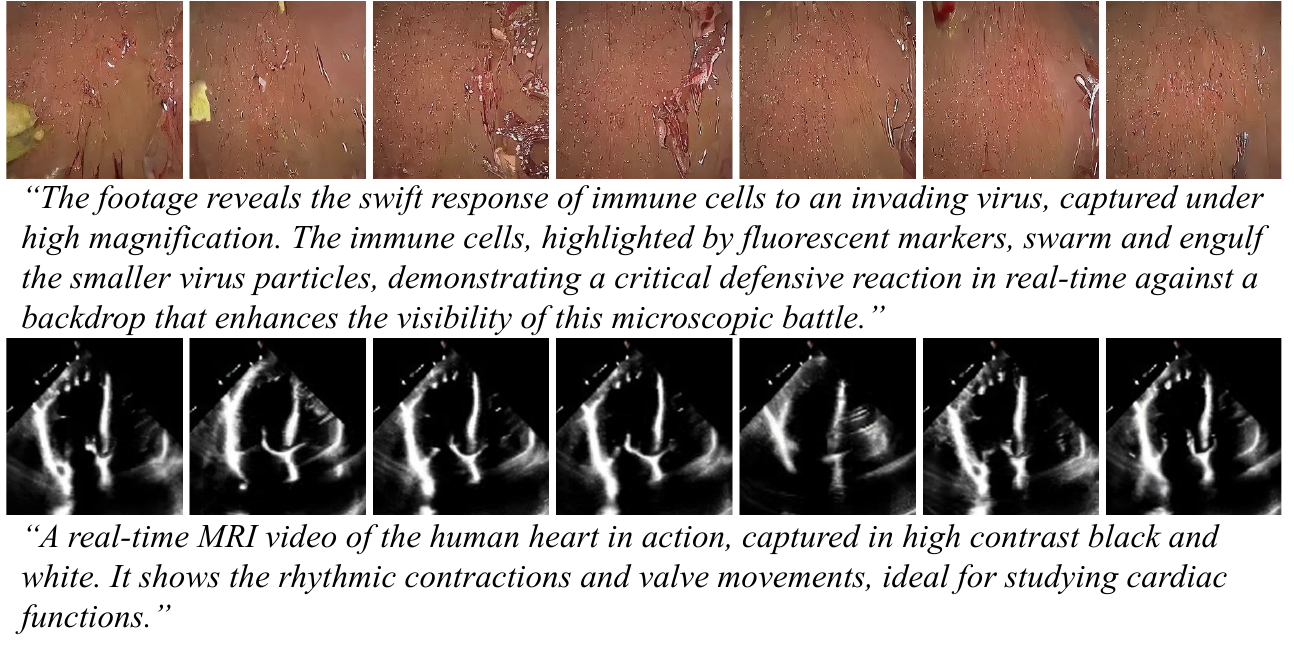}
    \caption{Two failure examples generated by our Bora and their corresponding medical prompts.}
    \label{fig: fail}
\end{figure*}

\noindent

\end{document}

%% file: weixiang/tab/data.tex
\begin{table*}[!t]
    \centering
    \resizebox{\linewidth}{!}{
    \begin{tabular}{cccccc}
    \toprule
       Modality  & Data Source & Origin Resolution & Length & Origin Size & Processed Size \\
    \midrule
    \multirow{3}{*}{Endoscopy} & Colonoscopy~\cite{mesejo2016computer} & $768\times5761$ & 10s+ & 76 & 210 \\
    \cline{2-6} 
      & Kvasir-Capsule~\cite{smedsrud2021kvasir} & $256\times448$ & / & 50 & 1000 \\
    \cline{2-6}
      & CholecTriplet~\cite{nwoye2022rendezvous} & $720\times576$ & 10s+ & 374 & 580 \\
    \midrule

    \multirow{2}{*}{Ultrasound} & Echo-Dynamic~\cite{ouyang2020video} & $112\times112$ & 5s & 10,030 & 10,030\\
    \cline{2-6} 
      & ULiver~\cite{de2013learning,petrusca2013hybrid} & $500\times480$ & 10s+ & 7 & 28 \\
    \midrule
    RT-MRI & 2drt~\cite{lim2021multispeaker} & $84\times84$ & 10s+ &  / & 1682 \\
    \midrule
    \multirow{3}{*}{Cell} & Tryp~\cite{anzaku2023tryp} & $1360\times1024$ & 10s+ & 114 & 188\\
    \cline{2-6}
      & CMTC-v1~\cite{anjum2020ctmc} & $320\times400$ & 10s+ & 86 & 258\\
    \cline{2-6}
      & VISEM~\cite{haugen2019visem} & $640\times480$ & 10s+ & 85 & 339\\
    \bottomrule
    \end{tabular}}
    \caption{Sources and detailed information of the data in our text-video pair dataset.}
    \label{tab: data}
\end{table*}

%% file: weixiang/dataset.tex
\section{Biomedical Text-Video Pair Dataset}

\label{sec: data}
Previously, there had been no exploration in the field of text-to-video generation within the biomedical domain. Consequently, there are no readily available biomedical text-video pair datasets. To address this gap, we leveraged the capabilities of LLM to create the first biomedical text-video pair dataset including four major biomedical modalities.

\subsection{Included Videos}

Our video data encompasses four primary biomedical modalities: endoscopic imaging, ultrasonography, real-time MRI, and cellular motility. Table A details the specific dataset sources along with their fundamental information. For varying resolutions, we standardize each to 256x256 pixels to facilitate model training. Regarding temporal length, if the original dataset's video duration exceeds ten seconds, it is uniformly recorded as 10s+. For videos that are excessively long, we determine a threshold \( K \) based on the degree of frame-to-frame variation. Sampling begins at zero frame intervals and progressively increases until the average inter-frame interval of the resultant video exceeds the predetermined threshold.

\subsection{LLMs Instruct Caption Generation}
Currently, a vast array of multimodal large language models (LLMs) support various input modes, including functioning as unimodal language models for processing purely textual inputs, accepting single image-text pairs or single images alone, as well as handling interleaved image-text pairs or multiple images, all with good performance. Furthermore,~\cite{llavamed,peng2023instruction} has demonstrated that data generated by LLMs can serve as high-quality training data. More importantly, LLMs that exhibit strong performance in general tasks, even without specific fine-tuning for the biomedical domain, still show excellent capabilities, such as the zero-shot performance of GPT-4 in the biomedical field~\cite{yan2023multimodal}. Therefore, we further explore the powerful capabilities of LLMs in the domain of biomedical videos, which involve temporal information, aiming to efficiently and accurately generate video descriptions.

In summary, we pre-process the video $X_{v}^i$ by evenly splitting it into $n$ frames $(\textit{f}_{1}, \cdots, \textit{f}_n)$ and sequentially transmit these images to LLM, obtaining descriptions $X_{desc}^i$. Then the origin video $X_{v}^i$ will be combined with its description $X_{desc}^i$ to form a text-video pair $X_i = (X_{v}^i, X_{desc}^i)$. However, during the process, we discovered that this straightforward approach tends to overlook the dataset's background, focusing primarily on describing the objects and movements within. To incorporate background information, we use an agent approach to transmit additional information to LLM, such as technical documents, research papers, or home pages related to the dataset. This not only enriches its biomedical background knowledge but also ensures that background information is not neglected, significantly enhancing the quality of the captions. More details about source data and processing can be found in Appendix~\ref{apx: data}.

%% file: weixiang/bora.tex
\section{Methods}
Following prior work~\cite{chen2023pixart,ramesh2022hierarchical,Feng_2023_CVPR,Wei_2023_ICCV,zhou2023shifted}, our architecture is divided into three modules: Text Encoder, Video Encoder, and Diffusion Block. Specifically, we initialize the weights using Open-Sora~\cite{opensora}, a framework capable of generating high-quality general video models, and subsequently conduct two-phase biomedical training on this basis.

\begin{figure}[!t]
    \centering
    \includegraphics[width = \linewidth]{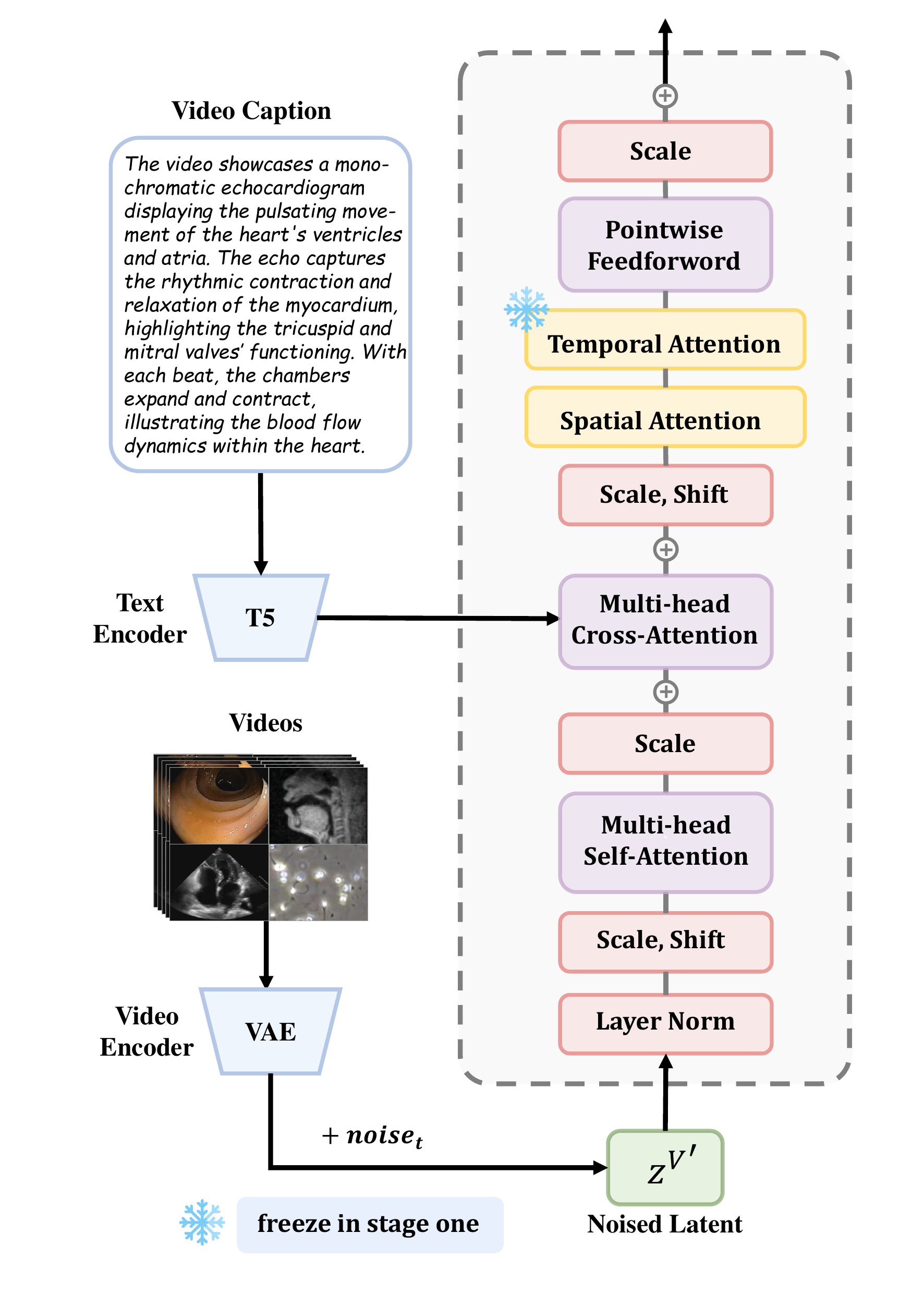}
    \caption{The overall architecture and training details of our Bora.}
    \label{fig: training}
\end{figure}

\subsection{Model Architecture}

\textbf{Text Encoder. } We adopt a pre-trained text encoder $T5$~\cite{raffel2020exploring} to encode the medical prompt. Specifically, we employ only the encoder portion of T5. It consists of multiple identical layers, each comprising a self-attention mechanism and a feedforward neural network. This architecture is capable of transforming the text \( X_{\text{desc}} \) into an representation \( z^T \) of medical prompts via the process \( z^T = \text{FFN}(\text{SelfAttention}(X_{desc})) \).

\noindent\textbf{Video Encoder. }
We compress the training data into a smaller representation within the latent space using a video encoder, which is then utilized for training subsequent blocks. For each video \(X_v\), we sample \(T\) frames. For each batch \(B_v\), we form a combination shaped as \((B, C, T)\), where \(B\) is the batch size, \(C\) is the number of channels, and \(T\) is the length of the time series (i.e., the number of sampled frames). We rearrange the sequences in the input data into a single frame, reshaping into \(B_v' = (B \times T, C)\). Using the mean \(\mu\) and variance \(\log \sigma^2\) output from a pre-trained VAE, we obtain a Gaussian distribution in the latent space \(q(z \mid x) = \mathcal{N}(z; \mu(x), \sigma^2(x))\). Sampling \(z^V\) from this distribution, we employ the reparameterization trick \(z^V = (\mu(x) + \sigma(x) \odot \epsilon) \times 0.18215\), where \(\epsilon \sim \mathcal{N}(0, I)\), allowing gradients to propagate through the stochastic sampling, and apply a scaling transformation. Finally, we rearrange the data back into the shape \((B, C, T)\) to match subsequent dimensions.

\noindent\textbf{Difusion Transformer. }
Diffusion models typically include a "forward process" and a "reverse process". The forward process incrementally adds noise to a data point \( x_0 \), transforming it completely into Gaussian noise \( x_T \). This process can be described as:
\begin{equation}
    q\left(x_t \mid x_{t-1}\right)=\mathcal{N}\left(x_t ; \sqrt{1-\beta_t} x_{t-1}, \beta_t I\right)
\end{equation}

Conversely, the reverse process attempts to recover the original data point from this noisy state. In conditions where text is present, the core approach involves combining text embeddings with diffusion states to modulate the reverse process as follows:
\begin{equation}
    p_\theta\left(x_{t-1} \mid x_t, c\right)=\mathcal{N}\left(x_{t-1} ; \mu_\theta\left(x_t, t, c\right), \sigma_t^2 I\right)
\end{equation}


\subsection{Adapting to Biomedical Domain}
To efficiently and effectively obtain a high-quality model for generating biomedical videos, we divide the training into two stages: Modal Alignment and Full-parameter Training.

\noindent\textbf{Biomedical Modal Alignment. }Although Open-Sora has not publicized the text-video pair data they created, our inference based on the source data suggests that it lacks any inclusion of medical concepts or scenarios. Furthermore, their parameters initiate from PixArt-$\alpha$~\cite{chen2023pixart}, which also lacks effective medical knowledge injection. Therefore, modal alignment is particularly important. To enhance efficiency in this step, we simplify the data and freeze a portion of the parameters. Regarding the data, for modalities with fewer total videos, we extract more frames randomly from each video, while for modalities with more videos, fewer frames are extracted from each to ensure performance balance across different modalities. Moreover, we modify the corresponding captions to \textit{"This is a {modal} video."} On the model front, we freeze the temporal attention to accelerate training. By employing these strategies, we efficiently provide simple guidance for video generation models that do not contain biomedical knowledge, laying a foundation for further training.

\noindent\textbf{Instruction Tuning. }In the second step, we first unfreeze the temporal attention module and then update the weights obtained from the previous step. During this stage, we train using the biomedical text-video pairs introduced in Sec~\ref{sec: data}, constructing a biomedical video generator. In fact, to ensure balance across different modalities, we did not utilize all the completed data sets. We used the modality with the smallest amount of data as a benchmark, balancing the quantities among the different modalities to prevent the occurrence of inter-modal chaos. For videos of varying lengths, we adopt different sampling intervals to capture richer temporal information. Experiments will demonstrate that our model not only possesses robust command-following capabilities, accurately transforming medical terminologies into corresponding videos but also ensures sufficient video quality.

%% file: weixiang/exp.tex
\section{Experiments}

\begin{figure*}[!t]
    \centering
    \includegraphics[width = \linewidth]{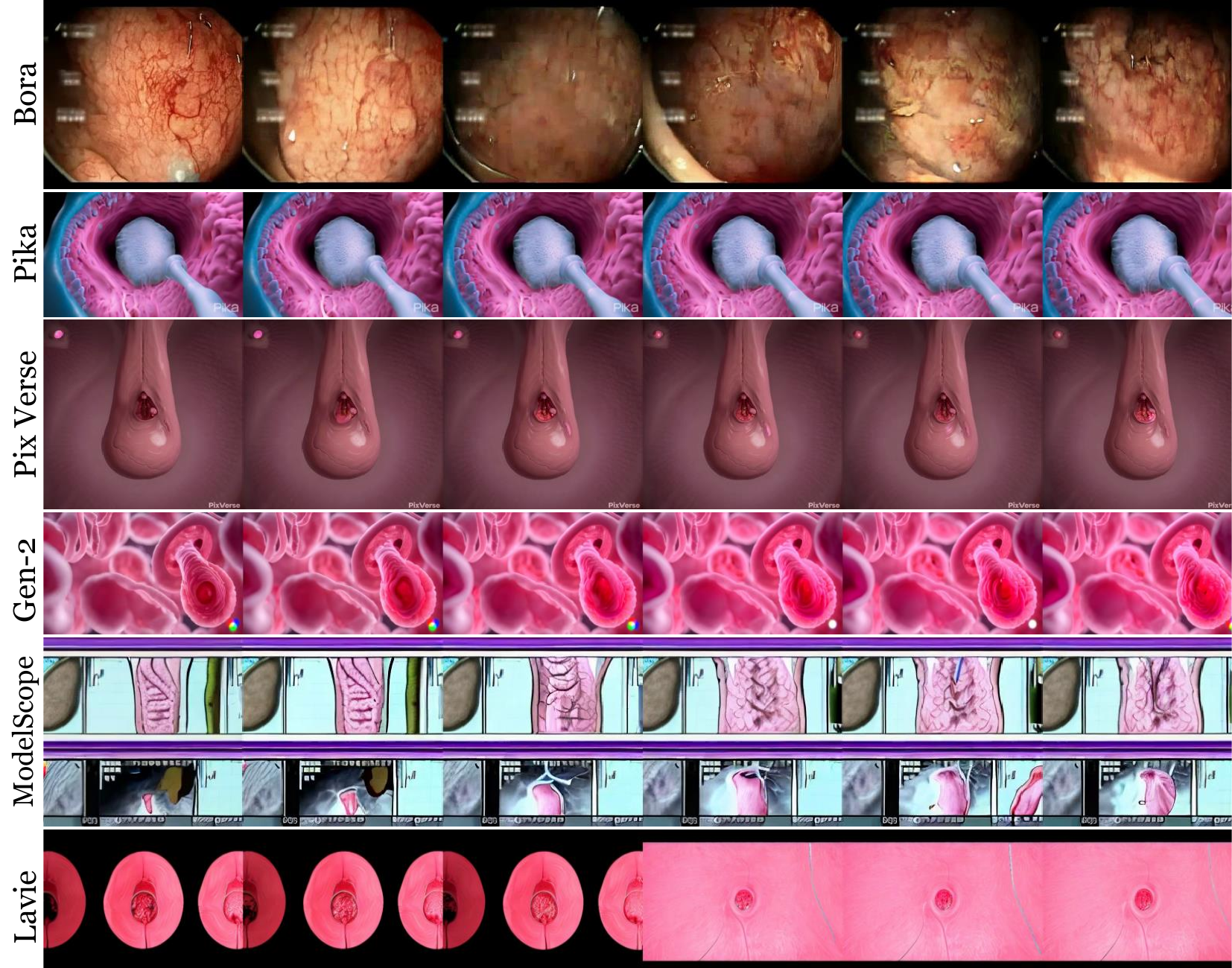}
    \caption{The comparison of generated video under the same prompt in endoscopy modal. From top to bottom are from Bora, Pika, PixVerse, Gen-2, ModelScope, and Lavie.}
    \label{fig: compare}
\end{figure*}

\subsection{Setup}

\noindent\textbf{Baseline Models. }At present, there are numerous available open-source or commercial video generation models. We have selected several high-performance models as benchmark models, including Pika~\cite{pika}, PixVerse~\cite{PixVerse}, Gen-2~\cite{Gen-2}, ModelScope~\cite{modelscope}, and Lavie~\cite{wang2023lavie}.

\noindent\textbf{Implemention Details. }For text-to-video generation, we employ GPT-4 to generate text prompts across different modalities. Specifically, we provide an overview of the desired modalities and some examples to learn from, and then prompt it to generate a certain number of text prompts. For certain models that have length restrictions on text prompts, we utilize GPT-4 to rewrite overly long prompts without altering their meaning. All generated text prompts are subsequently input into text-to-video models to produce videos.

\noindent All experiments are conducted on one TESLA A100 GPU, equipped with a substantial $80 \mathrm{~GB}$ of VRAM. The central processing was handled by $4 \times$ Intell(R) Xeon(R) Platinum 8362 56-Core Processors. The software environment was standardized on PyTorch version 2.2.2 and CUDA 12.1 for video generation and PyTorch version 1.13.1 and CUDA 12.1 for video evaluation. More details about training and evaluation can be found in Appendix~\ref{apx: impl}.

\noindent\textbf{Biomedical Instruction Following Metrics. }
For the biomedical field, our primary concern is to evaluate the authenticity of the generated videos and their understanding of biomedical information. Based on this, we have designed the following three metrics for evaluation.

\noindent\ding{182}Realism Rate Metrics: We use Video-Llama~\cite{videollama} to determine whether a video is from the real world. If it determines that the video is from the real world, it scores 1 point; otherwise, it scores 0 points. During the evaluation process, we found that sometimes it returns ambiguous answers such as \textit{"I am not sure if this is a real-world video."} To ensure the valid participation of all samples, we assign a score of 0.5 in such cases. The final realism rate is calculated as the average score across all samples.

\vspace{-0.6cm}
\begin{equation}
\begin{split}
I(x)&=VideoLlama(video) \\
Realism\ Rate&=\begin{cases}
 1 & if \ I(X)=Yes\\ 
 0 & if \ I(X)=Yes \\
0.5 & if \ I(X)=NotSure
\end{cases}
\end{split}
\end{equation}

\noindent\ding{183}Biomedical Understanding (BmU): Frankly, even if a video is judged to be completely virtual, it is still an excellent model if it can accurately convey the intent of the biomedical prompt. To this end, we propose Biomedical Understanding, a metric designed to evaluate the degree of adherence to prompts within the latent space. Specifically, we achieve this by having a Large Language Model (LLM) describe the video, then inputting the newly obtained description $\textit{T}_{new}$ text and the original prompt $\textit{T}_{ori}$ into BERT to calculate their pre-defined similarity in the embedding space. We obtain the new video descriptions using two methods: 1) passing image sequences as video (BmU-I), and 2) using Video-Llama~\cite{videollama} to describe the video (BmU-V). These two methods provide video descriptions from different perspectives, ensuring the accuracy of the evaluation.

\begin{equation}
BmU = cos(BERT(T_{new}^{(i)}), BERT(T_{ori}^{(i)}))
\end{equation}

\begin{table*}[!t]
    \centering
    \caption{The comparison of biomedical instruction following ability between other models. All the results are under biomedical prompts.}
    \resizebox{\linewidth}{!}{
    \begin{tabular}{c|cccc|c}
    \toprule
         Model & Realism Rate & \makecell[c]{BmU-I} & \makecell[c]{BmU-V} & BmU-ave & \makecell[c]{Default \\ Length(\textbf{s}) } \\
    \hline
         Pika~\cite{pika} & 0.14 & 0.60 & 0.67 &0.64 & 4\\
         PixVerse~\cite{PixVerse} & 0.06 & 0.71 & 0.69 &0.70 &4\\
         Gen-2~\cite{Gen-2} & 0.11 & 0.55 & 0.59 & 0.57&4\\
         ModelScope~\cite{modelscope} & 0.32 & 0.39 & 0.52 & 0.46&2 \\
         Lavie~\cite{wang2023lavie} & 0.20 & 0.47 & 0.53 &0.50 &2\\
         \midrule
         \textbf{Bora} & \textbf{0.66} & \textbf{0.83} & \textbf{0.89} & \textbf{0.86} & \textbf{5} \\
        \bottomrule
    \end{tabular}}
    \label{tab: following ability}
\end{table*}

\input{weixiang/tab/eva}

         
\vspace{-0.8cm}
\noindent\textbf{Video Quality Metrics. }We evaluate the comprehensive video quality of four covered biomedical modalities, following some of the basic metrics proposed in~\cite{Huang2023VBenchCB}. As our model is specialized for the biomedical domain, we have omitted the aesthetic scoring.

\noindent\ding{182} Subject Consistency, computed by the DINO~\cite{DINO} feature similarity across frames to assess whether the appearance is consistent throughout the video; 
\ding{183} Background Consistency, calculated by CLIP \cite{Radford2021LearningTV} feature similarity across frames to evaluate temporal consistency of the background;
\ding{184} Temporal Flicking, by selecting keyframes and calculating their average absolute deviation to access temporal consistency at details.
\ding{185} Motion Smoothness, which utilizes the motion priors in the video frame interpolation model AMT~\cite{amt} focuses on "move" to evaluate the smoothness of generated motions; 
\ding{186} Dynamic Degree, computed by employing RAFT \cite{raft} to see whether it contains large motions;
\ding{187} Imaging Quality, calculated by using MUSIQ \cite{musiq} image quality predictor;
\ding{188}Temporal style is determined by using ViCLIP~\cite{VICLIP} to compute the similarity between video features and temporal style descriptions, reflecting the style's consistency.

\subsection{Results}

The evaluation results of biomedical instruction following ability are shown in Table~\ref{tab: following ability}. It can be observed that our Bora significantly outperforms other models across all metrics. It should be noted that BERT~\cite{bert} maps text into an embedding space and performs comparisons based on vectors, it does not operate based on actual semantics. The BmU (Biomedical understanding) of other models is already at a considerably low level, resulting in the generated videos being almost entirely different. The comparison of video between Bora and other models is shown in Figure~\ref{fig: compare}.

Due to the significant discrepancies in the evaluation results from the previous phase, it is deemed unnecessary to conduct further comprehensive quality assessments on videos generated by other models. On one hand, comparing models that do not specialize in the biomedical domain is not fair; on the other hand, the biomedical videos they produce do not accurately reflect their true performance. Therefore, we shift our focus to evaluating videos generated across the four modalities we cover. For a concise and intuitive comparison of data, we still incorporate the evaluation results of Sora~\cite{yuan2024mora}. The results of comprehensive video quality are illustrated in Table~\ref{tab: quality}.

The cell is clearly ahead in terms of background consistency because its background is uniformly gray or bright on the slide. Due to the processed videos~\cite{endora}, endoscopy exhibits significant frame-to-frame variation, resulting in high dynamic degrees. Moreover, other metrics show little variation across the four modalities and generally perform well. The Image Quality of the Endoscope even slightly surpasses that of Sora. Despite some performance discrepancies between different modalities, the average scores still demonstrate its performance approaching that of Sora.

%% file: weixiang/tab/eva.tex
\vspace{1cm}

\begin{table*}[!t]
    \centering
        \caption{The comprehensive video quality and evaluation scores of four biomedical modalities videos generated by our \textbf{Bora}. \textit{Note: The results of Sora are only for comparison, not under biomedical prompts.}}
    \resizebox{\linewidth}{!}{
    \begin{tabular}{l|cccccc|c}
        \toprule
        Modality & \makecell[c]{Subject \\ Consistency} & \makecell[c]{Background \\ Consistency} & \makecell[c]{Temporal\\Flickering} & \makecell[c]{Motion \\ Smothness} & \makecell[c]{Dynamic \\ Dgree} & \makecell[c]{Imaging \\ Quality} & \makecell[c]{Temporal \\Style}  \\
        \hline 
        Endoscope & 0.89 & 0.96 & 0.90 & 0.91 & \textbf{0.90} & \textbf{0.60} & 0.23 \\
        Ultrasound & 0.91 & 0.94 & 0.97 & 0.98 & 0.29 & 0.37 & 0.21 \\
        RT-MRI & 0.90 & 0.96 & \textbf{0.99} & 0.99 & 0.28 & 0.19 & 0.23 \\
        Cell & 0.92 & \textbf{0.99} & 0.97 & 0.98 & 0.35 & 0.48 & 0.19  \\
        \hline
        Ave Score & 0.91 & 0.96 & 0.96 & 0.97 & 0.46 & 0.41 & 0.22 \\
        
    \bottomrule
    \toprule
    Sora & \textbf{0.95} & 0.96 & - & \textbf{1.0} & 0.69 & 0.58 & \textbf{0.35} \\
    \bottomrule
    \end{tabular}
    }
    \label{tab: quality}
\end{table*}

%% file: weixiang/limitation.tex
\section{Limitations}

\subsection{Highly Data-centric}

The collection and legal use of video data are often hindered by copyright protections, with these challenges becoming even more pronounced in the field of biomedical video. Beyond copyright, concerns surrounding privacy and ethics must be considered. High-quality biomedical videos are typically sourced from educational content at universities and institutions, where external access is restricted. This limitation forces reliance on open-source data for generating models across several biomedical modalities. Additionally, the execution of procedures in these videos demands high clarity, but biomedical processes, such as those using endoscopes, often produce videos of lower resolution. This underscores the importance of high-quality data for training biomedical generation models.

\subsection{Variable Quality of Captions}
Although numerous multimodal large language models (LLMs) can describe visual inputs, their performance in the biomedical domain significantly lags behind their capabilities in general domains. While some models are specifically fine-tuned for certain areas within the biomedical field, they are typically optimized for those particular domains and fail to effectively generate captions for other biomedical modalities. Moreover, a homogenization issue exists among the captions. Specifically, due to weak recognition capabilities regarding genuine medical details, the generated captions often repetitively echo similar content. This leads to confusion between different modalities, as demonstrated in Figure~\ref{fig: fail}, where descriptions intended for cell and RT-MRI scenarios result in endoscopic and ultrasound. The most accurate captions usually come from medical diagnoses or narratives by researchers, which not only raises the cost of generating captions but also poses potential risks to patient privacy. Striking a balance between accurate captions, manageable costs, and privacy regulations is crucial.

\subsection{Insufficiencies in Quality and Duration}
Despite our Bora model's ability to generate up to 5-second videos across various biomedical modalities, it underperforms when dealing with complex procedures or longer video durations. When we attempted to produce videos up to 16 seconds in length, there was a noticeable degradation in quality. This issue stems partly from a lack of high-quality, long-duration biomedical video data available for training, and partly from the suboptimal performance of our chosen base model in handling spatiotemporal interactions. In contrast, the best generators for regular videos, such as Sora, can produce high-quality videos exceeding one minute. Currently, we can only guarantee the quality of 5-second videos at a resolution of 256x256. This limitation urges us to further expand on spatiotemporal capabilities in future versions of our model.

%% file: main.bbl
\begin{thebibliography}{60}
\expandafter\ifx\csname natexlab\endcsname\relax\def\natexlab#1{#1}\fi

\bibitem[{Achiam et~al.(2023)Achiam, Adler, Agarwal, Ahmad, Akkaya, Aleman, Almeida, Altenschmidt, Altman, Anadkat et~al.}]{achiam2023gpt}
Josh Achiam, Steven Adler, Sandhini Agarwal, Lama Ahmad, Ilge Akkaya, Florencia~Leoni Aleman, Diogo Almeida, Janko Altenschmidt, Sam Altman, Shyamal Anadkat, et~al. 2023.
\newblock Gpt-4 technical report.
\newblock \emph{arXiv preprint arXiv:2303.08774}.

\bibitem[{Anjum and Gurari(2020)}]{anjum2020ctmc}
Samreen Anjum and Danna Gurari. 2020.
\newblock Ctmc: Cell tracking with mitosis detection dataset challenge.
\newblock In \emph{Proceedings of the IEEE/CVF Conference on Computer Vision and Pattern Recognition Workshops}, pages 982--983.

\bibitem[{Anzaku et~al.(2023)Anzaku, Mohammed, Ozbulak, Won, Hong, Krishnamoorthy, Van~Hoecke, Magez, Van~Messem, and De~Neve}]{anzaku2023tryp}
Esla~Timothy Anzaku, Mohammed~Aliy Mohammed, Utku Ozbulak, Jongbum Won, Hyesoo Hong, Janarthanan Krishnamoorthy, Sofie Van~Hoecke, Stefan Magez, Arnout Van~Messem, and Wesley De~Neve. 2023.
\newblock Tryp: a dataset of microscopy images of unstained thick blood smears for trypanosome detection.
\newblock \emph{Scientific Data}, 10(1):716.

\bibitem[{Bar-Tal et~al.(2024)Bar-Tal, Chefer, Tov, Herrmann, Paiss, Zada, Ephrat, Hur, Li, Michaeli et~al.}]{bar2024lumiere}
Omer Bar-Tal, Hila Chefer, Omer Tov, Charles Herrmann, Roni Paiss, Shiran Zada, Ariel Ephrat, Junhwa Hur, Yuanzhen Li, Tomer Michaeli, et~al. 2024.
\newblock Lumiere: A space-time diffusion model for video generation.
\newblock \emph{arXiv preprint arXiv:2401.12945}.

\bibitem[{Blattmann et~al.(2023{\natexlab{a}})Blattmann, Dockhorn, Kulal, Mendelevitch, Kilian, Lorenz, Levi, English, Voleti, Letts, Jampani, and Rombach}]{SVD}
Andreas Blattmann, Tim Dockhorn, Sumith Kulal, Daniel Mendelevitch, Maciej Kilian, Dominik Lorenz, Yam Levi, Zion English, Vikram Voleti, Adam Letts, Varun Jampani, and Robin Rombach. 2023{\natexlab{a}}.
\newblock \href {http://arxiv.org/abs/arXiv:2311.15127} {Stable video diffusion: Scaling latent video diffusion models to large datasets}.

\bibitem[{Blattmann et~al.(2023{\natexlab{b}})Blattmann, Dockhorn, Kulal, Mendelevitch, Kilian, Lorenz, Levi, English, Voleti, Letts et~al.}]{blattmann2023stable}
Andreas Blattmann, Tim Dockhorn, Sumith Kulal, Daniel Mendelevitch, Maciej Kilian, Dominik Lorenz, Yam Levi, Zion English, Vikram Voleti, Adam Letts, et~al. 2023{\natexlab{b}}.
\newblock Stable video diffusion: Scaling latent video diffusion models to large datasets.
\newblock \emph{arXiv preprint arXiv:2311.15127}.

\bibitem[{Caron et~al.(2021)Caron, Touvron, Misra, J{\'e}gou, Mairal, Bojanowski, and Joulin}]{DINO}
Mathilde Caron, Hugo Touvron, Ishan Misra, Herv{\'e} J{\'e}gou, Julien Mairal, Piotr Bojanowski, and Armand Joulin. 2021.
\newblock Emerging properties in self-supervised vision transformers.
\newblock In \emph{Proceedings of the IEEE/CVF international conference on computer vision}, pages 9650--9660.

\bibitem[{Chen et~al.(2023)Chen, Yu, Ge, Yao, Xie, Wu, Wang, Kwok, Luo, Lu et~al.}]{chen2023pixart}
Junsong Chen, Jincheng Yu, Chongjian Ge, Lewei Yao, Enze Xie, Yue Wu, Zhongdao Wang, James Kwok, Ping Luo, Huchuan Lu, et~al. 2023.
\newblock Pixart-$\alpha $: Fast training of diffusion transformer for photorealistic text-to-image synthesis.
\newblock \emph{arXiv preprint arXiv:2310.00426}.

\bibitem[{Chu et~al.(2020)Chu, Xie, Mayer, Leal-Taixé, and Thuerey}]{Chu_2020}
Mengyu Chu, You Xie, Jonas Mayer, Laura Leal-Taixé, and Nils Thuerey. 2020.
\newblock \href {https://doi.org/10.1145/3386569.3392457} {Learning temporal coherence via self-supervision for gan-based video generation}.
\newblock \emph{ACM Transactions on Graphics}, 39(4).

\bibitem[{Croitoru et~al.(2023)Croitoru, Hondru, Ionescu, and Shah}]{croitoru2023diffusion}
Florinel-Alin Croitoru, Vlad Hondru, Radu~Tudor Ionescu, and Mubarak Shah. 2023.
\newblock Diffusion models in vision: A survey.
\newblock \emph{IEEE Transactions on Pattern Analysis and Machine Intelligence}.

\bibitem[{De~Luca et~al.(2013)De~Luca, Tschannen, Sz{\'e}kely, and Tanner}]{de2013learning}
Valeria De~Luca, Michael Tschannen, G{\'a}bor Sz{\'e}kely, and Christine Tanner. 2013.
\newblock A learning-based approach for fast and robust vessel tracking in long ultrasound sequences.
\newblock In \emph{Medical Image Computing and Computer-Assisted Intervention--MICCAI 2013: 16th International Conference, Nagoya, Japan, September 22-26, 2013, Proceedings, Part I 16}, pages 518--525. Springer.

\bibitem[{Devlin et~al.(2018)Devlin, Chang, Lee, and Toutanova}]{bert}
Jacob Devlin, Ming-Wei Chang, Kenton Lee, and Kristina Toutanova. 2018.
\newblock Bert: Pre-training of deep bidirectional transformers for language understanding.
\newblock \emph{arXiv preprint arXiv:1810.04805}.

\bibitem[{Dorjsembe et~al.(2022)Dorjsembe, Odonchimed, and Xiao}]{dorjsembe2022three}
Zolnamar Dorjsembe, Sodtavilan Odonchimed, and Furen Xiao. 2022.
\newblock Three-dimensional medical image synthesis with denoising diffusion probabilistic models.
\newblock In \emph{Medical Imaging with Deep Learning}.

\bibitem[{Feng et~al.(2023)Feng, Zhang, Yu, Fang, Li, Chen, Lu, Liu, Yin, Feng, Sun, Chen, Tian, Wu, and Wang}]{Feng_2023_CVPR}
Zhida Feng, Zhenyu Zhang, Xintong Yu, Yewei Fang, Lanxin Li, Xuyi Chen, Yuxiang Lu, Jiaxiang Liu, Weichong Yin, Shikun Feng, Yu~Sun, Li~Chen, Hao Tian, Hua Wu, and Haifeng Wang. 2023.
\newblock Ernie-vilg 2.0: Improving text-to-image diffusion model with knowledge-enhanced mixture-of-denoising-experts.
\newblock In \emph{Proceedings of the IEEE/CVF Conference on Computer Vision and Pattern Recognition (CVPR)}, pages 10135--10145.

\bibitem[{Ge et~al.(2023)Ge, Nah, Liu, Poon, Tao, Catanzaro, Jacobs, Huang, Liu, and Balaji}]{ge2023preserve}
Songwei Ge, Seungjun Nah, Guilin Liu, Tyler Poon, Andrew Tao, Bryan Catanzaro, David Jacobs, Jia-Bin Huang, Ming-Yu Liu, and Yogesh Balaji. 2023.
\newblock Preserve your own correlation: A noise prior for video diffusion models.
\newblock In \emph{Proceedings of the IEEE/CVF International Conference on Computer Vision}, pages 22930--22941.

\bibitem[{Guo et~al.(2023)Guo, Yang, Rao, Wang, Qiao, Lin, and Dai}]{guo2023animatediff}
Yuwei Guo, Ceyuan Yang, Anyi Rao, Yaohui Wang, Yu~Qiao, Dahua Lin, and Bo~Dai. 2023.
\newblock Animatediff: Animate your personalized text-to-image diffusion models without specific tuning.
\newblock \emph{arXiv preprint arXiv:2307.04725}.

\bibitem[{Haugen et~al.(2019)Haugen, Hicks, Andersen, Witczak, Hammer, Borgli, Halvorsen, and Riegler}]{haugen2019visem}
Trine~B Haugen, Steven~A Hicks, Jorunn~M Andersen, Oliwia Witczak, Hugo~L Hammer, Rune Borgli, P{\aa}l Halvorsen, and Michael Riegler. 2019.
\newblock Visem: A multimodal video dataset of human spermatozoa.
\newblock In \emph{Proceedings of the 10th ACM Multimedia Systems Conference}, pages 261--266.

\bibitem[{Ho et~al.(2022)Ho, Chan, Saharia, Whang, Gao, Gritsenko, Kingma, Poole, Norouzi, Fleet et~al.}]{ho2022imagen}
Jonathan Ho, William Chan, Chitwan Saharia, Jay Whang, Ruiqi Gao, Alexey Gritsenko, Diederik~P Kingma, Ben Poole, Mohammad Norouzi, David~J Fleet, et~al. 2022.
\newblock Imagen video: High definition video generation with diffusion models.
\newblock \emph{arXiv preprint arXiv:2210.02303}.

\bibitem[{Ho and Salimans(2022)}]{ho2022classifier}
Jonathan Ho and Tim Salimans. 2022.
\newblock Classifier-free diffusion guidance.
\newblock \emph{arXiv preprint arXiv:2207.12598}.

\bibitem[{Huang et~al.(2023)Huang, He, Yu, Zhang, Si, Jiang, Zhang, Wu, Jin, Chanpaisit, Wang, Chen, Wang, Lin, Qiao, and Liu}]{Huang2023VBenchCB}
Ziqi Huang, Yinan He, Jiashuo Yu, Fan Zhang, Chenyang Si, Yuming Jiang, Yuanhan Zhang, Tianxing Wu, Qingyang Jin, Nattapol Chanpaisit, Yaohui Wang, Xinyuan Chen, Limin Wang, Dahua Lin, Yu~Qiao, and Ziwei Liu. 2023.
\newblock \href {https://api.semanticscholar.org/CorpusID:265506207} {Vbench: Comprehensive benchmark suite for video generative models}.
\newblock \emph{ArXiv}, abs/2311.17982.

\bibitem[{Ke et~al.(2021)Ke, Wang, Wang, Milanfar, and Yang}]{musiq}
Junjie Ke, Qifei Wang, Yilin Wang, Peyman Milanfar, and Feng Yang. 2021.
\newblock Musiq: Multi-scale image quality transformer.
\newblock In \emph{Proceedings of the IEEE/CVF international conference on computer vision}, pages 5148--5157.

\bibitem[{Li et~al.(2024{\natexlab{a}})Li, Liu, Liu, Feng, Li, Liu, Chen, Shao, and Yuan}]{endora}
Chenxin Li, Hengyu Liu, Yifan Liu, Brandon~Y. Feng, Wuyang Li, Xinyu Liu, Zhen Chen, Jing Shao, and Yixuan Yuan. 2024{\natexlab{a}}.
\newblock \href {http://arxiv.org/abs/arXiv:2403.11050} {Endora: Video generation models as endoscopy simulators}.

\bibitem[{Li et~al.(2024{\natexlab{b}})Li, Wong, Zhang, Usuyama, Liu, Yang, Naumann, Poon, and Gao}]{llavamed}
Chunyuan Li, Cliff Wong, Sheng Zhang, Naoto Usuyama, Haotian Liu, Jianwei Yang, Tristan Naumann, Hoifung Poon, and Jianfeng Gao. 2024{\natexlab{b}}.
\newblock Llava-med: Training a large language-and-vision assistant for biomedicine in one day.
\newblock \emph{Advances in Neural Information Processing Systems}, 36.

\bibitem[{Li et~al.(2017)Li, Min, Shen, Carlson, and Carin}]{li2017video}
Yitong Li, Martin~Renqiang Min, Dinghan Shen, David Carlson, and Lawrence Carin. 2017.
\newblock \href {http://arxiv.org/abs/1710.00421} {Video generation from text}.

\bibitem[{Li et~al.(2023)Li, Zhu, Han, Hou, Guo, and Cheng}]{amt}
Zhen Li, Zuo-Liang Zhu, Ling-Hao Han, Qibin Hou, Chun-Le Guo, and Ming-Ming Cheng. 2023.
\newblock Amt: All-pairs multi-field transforms for efficient frame interpolation.
\newblock In \emph{Proceedings of the IEEE/CVF Conference on Computer Vision and Pattern Recognition}, pages 9801--9810.

\bibitem[{Lim et~al.(2021)Lim, Toutios, Bliesener, Tian, Lingala, Vaz, Sorensen, Oh, Harper, Chen et~al.}]{lim2021multispeaker}
Yongwan Lim, Asterios Toutios, Yannick Bliesener, Ye~Tian, Sajan~Goud Lingala, Colin Vaz, Tanner Sorensen, Miran Oh, Sarah Harper, Weiyi Chen, et~al. 2021.
\newblock A multispeaker dataset of raw and reconstructed speech production real-time mri video and 3d volumetric images.
\newblock \emph{Scientific data}, 8(1):187.

\bibitem[{Liu et~al.(2024)Liu, Zhang, Li, Yan, Gao, Chen, Yuan, Huang, Sun, Gao et~al.}]{liu2024sora}
Yixin Liu, Kai Zhang, Yuan Li, Zhiling Yan, Chujie Gao, Ruoxi Chen, Zhengqing Yuan, Yue Huang, Hanchi Sun, Jianfeng Gao, et~al. 2024.
\newblock Sora: A review on background, technology, limitations, and opportunities of large vision models.
\newblock \emph{arXiv preprint arXiv:2402.17177}.

\bibitem[{Ma et~al.(2024)Ma, Wang, Jia, Chen, Liu, Li, Chen, and Qiao}]{ma2024latte}
Xin Ma, Yaohui Wang, Gengyun Jia, Xinyuan Chen, Ziwei Liu, Yuan-Fang Li, Cunjian Chen, and Yu~Qiao. 2024.
\newblock Latte: Latent diffusion transformer for video generation.
\newblock \emph{arXiv preprint arXiv:2401.03048}.

\bibitem[{Mesejo et~al.(2016)Mesejo, Pizarro, Abergel, Rouquette, Beorchia, Poincloux, and Bartoli}]{mesejo2016computer}
Pablo Mesejo, Daniel Pizarro, Armand Abergel, Olivier Rouquette, Sylvain Beorchia, Laurent Poincloux, and Adrien Bartoli. 2016.
\newblock Computer-aided classification of gastrointestinal lesions in regular colonoscopy.
\newblock \emph{IEEE transactions on medical imaging}, 35(9):2051--2063.

\bibitem[{ML(2023)}]{Gen-2}
Runway ML. 2023.
\newblock Text to video.
\newblock URL \url{https://runwayml.com/ai-tools/gen-2-text-to-video/}.

\bibitem[{Mogavi et~al.(2024)Mogavi, Wang, Tu, Hadan, Sgandurra, Hui, and Nacke}]{mogavi2024sora}
Reza~Hadi Mogavi, Derrick Wang, Joseph Tu, Hilda Hadan, Sabrina~A Sgandurra, Pan Hui, and Lennart~E Nacke. 2024.
\newblock Sora openai's prelude: Social media perspectives on sora openai and the future of ai video generation.
\newblock \emph{arXiv preprint arXiv:2403.14665}.

\bibitem[{Nwoye et~al.(2022)Nwoye, Yu, Gonzalez, Seeliger, Mascagni, Mutter, Marescaux, and Padoy}]{nwoye2022rendezvous}
Chinedu~Innocent Nwoye, Tong Yu, Cristians Gonzalez, Barbara Seeliger, Pietro Mascagni, Didier Mutter, Jacques Marescaux, and Nicolas Padoy. 2022.
\newblock Rendezvous: Attention mechanisms for the recognition of surgical action triplets in endoscopic videos.
\newblock \emph{Medical Image Analysis}, 78:102433.

\bibitem[{{OpenAI}(2024)}]{openai2024sora}
{OpenAI}. 2024.
\newblock Sora: Creating video from text.
\newblock \url{https://openai.com/sora}.

\bibitem[{OpenAI et~al.(2023)OpenAI, Achiam, Adler, Agarwal, Ahmad, Akkaya, Aleman, and Almeida}]{GPT4report}
OpenAI, Josh Achiam, Steven Adler, Sandhini Agarwal, Lama Ahmad, Ilge Akkaya, Florencia~Leoni Aleman, and Diogo Almeida. 2023.
\newblock \href {http://arxiv.org/abs/arXiv:2303.08774} {Gpt-4 technical report}.

\bibitem[{Ouyang et~al.(2020)Ouyang, He, Ghorbani, Yuan, Ebinger, Langlotz, Heidenreich, Harrington, Liang, Ashley et~al.}]{ouyang2020video}
David Ouyang, Bryan He, Amirata Ghorbani, Neal Yuan, Joseph Ebinger, Curtis~P Langlotz, Paul~A Heidenreich, Robert~A Harrington, David~H Liang, Euan~A Ashley, et~al. 2020.
\newblock Video-based ai for beat-to-beat assessment of cardiac function.
\newblock \emph{Nature}, 580(7802):252--256.

\bibitem[{Peng et~al.(2023)Peng, Li, He, Galley, and Gao}]{peng2023instruction}
Baolin Peng, Chunyuan Li, Pengcheng He, Michel Galley, and Jianfeng Gao. 2023.
\newblock Instruction tuning with gpt-4.
\newblock \emph{arXiv preprint arXiv:2304.03277}.

\bibitem[{Petrusca et~al.(2013)Petrusca, Cattin, De~Luca, Preiswerk, Celicanin, Auboiroux, Viallon, Arnold, Santini, Terraz et~al.}]{petrusca2013hybrid}
Lorena Petrusca, Philippe Cattin, Valeria De~Luca, Frank Preiswerk, Zarko Celicanin, Vincent Auboiroux, Magalie Viallon, Patrik Arnold, Francesco Santini, Sylvain Terraz, et~al. 2013.
\newblock Hybrid ultrasound/magnetic resonance simultaneous acquisition and image fusion for motion monitoring in the upper abdomen.
\newblock \emph{Investigative radiology}, 48(5):333--340.

\bibitem[{Pika(2023)}]{pika}
Pika. 2023.
\newblock Pika art.
\newblock URL \url{https://pika.art/home}.

\bibitem[{PixVerse(2024)}]{PixVerse}
PixVerse. 2024.
\newblock Pixverse.
\newblock URL \url{https://pixverse.ai/}.

\bibitem[{Radford et~al.(2021)Radford, Kim, Hallacy, Ramesh, Goh, Agarwal, Sastry, Askell, Mishkin, Clark, Krueger, and Sutskever}]{Radford2021LearningTV}
Alec Radford, Jong~Wook Kim, Chris Hallacy, A.~Ramesh, Gabriel Goh, Sandhini Agarwal, Girish Sastry, Amanda Askell, Pamela Mishkin, Jack Clark, Gretchen Krueger, and Ilya Sutskever. 2021.
\newblock Learning transferable visual models from natural language supervision.
\newblock In \emph{ICML}.

\bibitem[{Raffel et~al.(2020)Raffel, Shazeer, Roberts, Lee, Narang, Matena, Zhou, Li, and Liu}]{raffel2020exploring}
Colin Raffel, Noam Shazeer, Adam Roberts, Katherine Lee, Sharan Narang, Michael Matena, Yanqi Zhou, Wei Li, and Peter~J Liu. 2020.
\newblock Exploring the limits of transfer learning with a unified text-to-text transformer.
\newblock \emph{Journal of machine learning research}, 21(140):1--67.

\bibitem[{Ramesh et~al.(2022)Ramesh, Dhariwal, Nichol, Chu, and Chen}]{ramesh2022hierarchical}
Aditya Ramesh, Prafulla Dhariwal, Alex Nichol, Casey Chu, and Mark Chen. 2022.
\newblock Hierarchical text-conditional image generation with clip latents.
\newblock \emph{arXiv preprint arXiv:2204.06125}, 1(2):3.

\bibitem[{Rombach et~al.(2022)Rombach, Blattmann, Lorenz, Esser, and Ommer}]{rombach2022high}
Robin Rombach, Andreas Blattmann, Dominik Lorenz, Patrick Esser, and Bj{\"o}rn Ommer. 2022.
\newblock High-resolution image synthesis with latent diffusion models.
\newblock In \emph{Proceedings of the IEEE/CVF conference on computer vision and pattern recognition}, pages 10684--10695.

\bibitem[{Saharia et~al.(2022)Saharia, Chan, Saxena, Li, Whang, Denton, Ghasemipour, Gontijo~Lopes, Karagol~Ayan, Salimans et~al.}]{saharia2022photorealistic}
Chitwan Saharia, William Chan, Saurabh Saxena, Lala Li, Jay Whang, Emily~L Denton, Kamyar Ghasemipour, Raphael Gontijo~Lopes, Burcu Karagol~Ayan, Tim Salimans, et~al. 2022.
\newblock Photorealistic text-to-image diffusion models with deep language understanding.
\newblock \emph{Advances in neural information processing systems}, 35:36479--36494.

\bibitem[{Singer et~al.(2022)Singer, Polyak, Hayes, Yin, An, Zhang, Hu, Yang, Ashual, Gafni, Parikh, Gupta, and Taigman}]{singer2022makeavideo}
Uriel Singer, Adam Polyak, Thomas Hayes, Xi~Yin, Jie An, Songyang Zhang, Qiyuan Hu, Harry Yang, Oron Ashual, Oran Gafni, Devi Parikh, Sonal Gupta, and Yaniv Taigman. 2022.
\newblock \href {http://arxiv.org/abs/2209.14792} {Make-a-video: Text-to-video generation without text-video data}.

\bibitem[{Smedsrud et~al.(2021)Smedsrud, Thambawita, Hicks, Gjestang, Nedrejord, N{\ae}ss, Borgli, Jha, Berstad, Eskeland et~al.}]{smedsrud2021kvasir}
Pia~H Smedsrud, Vajira Thambawita, Steven~A Hicks, Henrik Gjestang, Oda~Olsen Nedrejord, Espen N{\ae}ss, Hanna Borgli, Debesh Jha, Tor Jan~Derek Berstad, Sigrun~L Eskeland, et~al. 2021.
\newblock Kvasir-capsule, a video capsule endoscopy dataset.
\newblock \emph{Scientific Data}, 8(1):142.

\bibitem[{Teed and Deng(2020)}]{raft}
Zachary Teed and Jia Deng. 2020.
\newblock Raft: Recurrent all-pairs field transforms for optical flow.
\newblock In \emph{Computer Vision--ECCV 2020: 16th European Conference, Glasgow, UK, August 23--28, 2020, Proceedings, Part II 16}, pages 402--419. Springer.

\bibitem[{Tulyakov et~al.(2017)Tulyakov, Liu, Yang, and Kautz}]{tulyakov2017mocogan}
Sergey Tulyakov, Ming-Yu Liu, Xiaodong Yang, and Jan Kautz. 2017.
\newblock \href {http://arxiv.org/abs/1707.04993} {Mocogan: Decomposing motion and content for video generation}.

\bibitem[{Wang et~al.(2023{\natexlab{a}})Wang, Yuan, Chen, Zhang, Wang, and Zhang}]{modelscope}
Jiuniu Wang, Hangjie Yuan, Dayou Chen, Yingya Zhang, Xiang Wang, and Shiwei Zhang. 2023{\natexlab{a}}.
\newblock Modelscope text-to-video technical report.
\newblock \emph{arXiv preprint arXiv:2308.06571}.

\bibitem[{Wang et~al.(2020)Wang, Bilinski, Bremond, and Dantcheva}]{yaohui2020imaginator}
Yaohui Wang, Piotr Bilinski, Francois Bremond, and Antitza Dantcheva. 2020.
\newblock \href {https://doi.org/10.1109/WACV45572.2020.9093492} {Imaginator: Conditional spatio-temporal gan for video generation}.
\newblock In \emph{2020 IEEE Winter Conference on Applications of Computer Vision (WACV)}, pages 1149--1158.

\bibitem[{Wang et~al.(2023{\natexlab{b}})Wang, Chen, Ma, Zhou, Huang, Wang, Yang, He, Yu, Yang et~al.}]{wang2023lavie}
Yaohui Wang, Xinyuan Chen, Xin Ma, Shangchen Zhou, Ziqi Huang, Yi~Wang, Ceyuan Yang, Yinan He, Jiashuo Yu, Peiqing Yang, et~al. 2023{\natexlab{b}}.
\newblock Lavie: High-quality video generation with cascaded latent diffusion models.
\newblock \emph{arXiv preprint arXiv:2309.15103}.

\bibitem[{Wang et~al.(2023{\natexlab{c}})Wang, He, Li, Li, Yu, Ma, Li, Chen, Chen, Wang et~al.}]{VICLIP}
Yi~Wang, Yinan He, Yizhuo Li, Kunchang Li, Jiashuo Yu, Xin Ma, Xinhao Li, Guo Chen, Xinyuan Chen, Yaohui Wang, et~al. 2023{\natexlab{c}}.
\newblock Internvid: A large-scale video-text dataset for multimodal understanding and generation.
\newblock \emph{arXiv preprint arXiv:2307.06942}.

\bibitem[{Wei et~al.(2023)Wei, Zhang, Ji, Bai, Zhang, and Zuo}]{Wei_2023_ICCV}
Yuxiang Wei, Yabo Zhang, Zhilong Ji, Jinfeng Bai, Lei Zhang, and Wangmeng Zuo. 2023.
\newblock Elite: Encoding visual concepts into textual embeddings for customized text-to-image generation.
\newblock In \emph{Proceedings of the IEEE/CVF International Conference on Computer Vision (ICCV)}, pages 15943--15953.

\bibitem[{Xu et~al.(2023)Xu, Sun, Peng, Visweswaran, and Batmanghelich}]{xu2023medsyn}
Yanwu Xu, Li~Sun, Wei Peng, Shyam Visweswaran, and Kayhan Batmanghelich. 2023.
\newblock \href {http://arxiv.org/abs/2310.03559} {Medsyn: Text-guided anatomy-aware synthesis of high-fidelity 3d ct images}.

\bibitem[{Yan et~al.(2023)Yan, Zhang, Zhou, He, Li, and Sun}]{yan2023multimodal}
Zhiling Yan, Kai Zhang, Rong Zhou, Lifang He, Xiang Li, and Lichao Sun. 2023.
\newblock Multimodal chatgpt for medical applications: an experimental study of gpt-4v.
\newblock \emph{arXiv preprint arXiv:2310.19061}.

\bibitem[{Yuan et~al.(2024)Yuan, Chen, Li, Jia, He, Wang, and Sun}]{yuan2024mora}
Zhengqing Yuan, Ruoxi Chen, Zhaoxu Li, Haolong Jia, Lifang He, Chi Wang, and Lichao Sun. 2024.
\newblock Mora: Enabling generalist video generation via a multi-agent framework.
\newblock \emph{arXiv preprint arXiv:2403.13248}.

\bibitem[{Zhang et~al.(2023)Zhang, Li, and Bing}]{videollama}
Hang Zhang, Xin Li, and Lidong Bing. 2023.
\newblock \href {https://arxiv.org/abs/2306.02858} {Video-llama: An instruction-tuned audio-visual language model for video understanding}.
\newblock \emph{arXiv preprint arXiv:2306.02858}.

\bibitem[{Zheng et~al.(2024)Zheng, Peng, and You}]{opensora}
Zangwei Zheng, Xiangyu Peng, and Yang You. 2024.
\newblock \href {https://github.com/hpcaitech/Open-Sora} {Open-sora: Democratizing efficient video production for all}.

\bibitem[{Zhou et~al.(2023{\natexlab{a}})Zhou, Wang, Yan, Lv, Zhu, and Feng}]{zhou2023magicvideo}
Daquan Zhou, Weimin Wang, Hanshu Yan, Weiwei Lv, Yizhe Zhu, and Jiashi Feng. 2023{\natexlab{a}}.
\newblock \href {http://arxiv.org/abs/2211.11018} {Magicvideo: Efficient video generation with latent diffusion models}.

\bibitem[{Zhou et~al.(2023{\natexlab{b}})Zhou, Liu, Zhu, Yang, Chen, and Xu}]{zhou2023shifted}
Yufan Zhou, Bingchen Liu, Yizhe Zhu, Xiao Yang, Changyou Chen, and Jinhui Xu. 2023{\natexlab{b}}.
\newblock \href {http://arxiv.org/abs/2211.15388} {Shifted diffusion for text-to-image generation}.

\end{thebibliography}
